\documentclass[twoside,journal]{IEEEtran}
\ifCLASSINFOpdf
\else
\fi
\usepackage[usenames, dvipsnames]{xcolor}
\usepackage[colorinlistoftodos]{todonotes}
\usepackage{setspace}
\usepackage{amsfonts}
\usepackage{cite}
\usepackage{hyperref}
\usepackage{tabularx}
\usepackage{subcaption}
\usepackage{booktabs}
\usepackage{array}
\usepackage{longtable}
\usepackage{pgfgantt}
\usepackage{amsmath}
\usepackage{notoccite}
\usepackage{enumitem}
\usepackage{tikz}
\usetikzlibrary{shapes.geometric, arrows}
\usepackage[export]{adjustbox}
\usepackage{multirow}
\usepackage{flushend}
\usepackage{soul}
\usepackage[switch]{lineno}

\newcommand{\eq}[1]   {Eq.\,(\ref{#1})}		

\newcommand{\fig}[1]  {Fig.\,\ref{#1}}		
\newcommand{\tab}[1]  {Table~\ref{#1}}		
\newcommand{\secn}[1] {Section~\ref{#1}}	
\newcommand{\ssec}[1] {Subsection~\ref{#1}}	

\newcommand{\revised}[3]{#2}

\begin{document}
%
\title{Parameter Optimization and Learning in a Spiking Neural Network for UAV Obstacle Avoidance targeting Neuromorphic Processors}
%
%
%
%

\author{Llewyn~Salt,
        David~Howard,~\IEEEmembership{Member,~IEEE,}
        Giacomo~Indiveri,~\IEEEmembership{Senior Member,~IEEE,}
        Yulia~Sandamirskaya,~\IEEEmembership{Member,~IEEE,}
\thanks{\IEEEcompsocthanksitem L. Salt is with the School
of Information Technology and Electrical Engineering, University of Queensland, Queensland,
Australia.}%
\thanks{D. Howard is with the Robotics and Autonomous Systems Group in the Cyberphysical System Program, CSIRO, Queensland, Australia.}%
\thanks{G. Indiveri and Y. Sandamirskaya are with the Institute of Neuroinformatics, University of Zurich and ETH Zurich, Zurich, Switzerland.
}%
}

%
%

\markboth{Journal of \LaTeX\ Class Files,~Vol.~14, No.~8, August~2015}%
{Shell \MakeLowercase{\textit{et al.}}: Bare Demo of IEEEtran.cls for IEEE Journals}
\maketitle

\begin{abstract}
  The Lobula Giant Movement Detector (LGMD) is an identified neuron of the locust that detects looming objects and triggers the insect's escape responses. Understanding the neural principles and network structure that lead to these fast and robust responses can facilitate the design of efficient obstacle avoidance strategies for robotic applications. Here we present a neuromorphic spiking neural network model of the LGMD driven by the output of a neuromorphic Dynamic Vision Sensor (DVS), which incorporates spiking frequency adaptation and synaptic plasticity mechanisms, and which can be mapped onto existing neuromorphic processor chips. However, as the model has a wide range of parameters, and the mixed signal analogue-digital circuits used to implement the model are affected by variability and noise, it is necessary to optimise the parameters to produce robust and reliable responses. Here we propose to use Differential Evolution (DE) and Bayesian Optimisation (BO) techniques to optimise the parameter space and investigate the use of Self-Adaptive Differential Evolution (SADE) to ameliorate the difficulties of finding appropriate input parameters for the DE technique. We quantify the performance of the methods proposed with a comprehensive comparison of different optimisers applied to the model, and demonstrate the validity of the approach proposed using recordings made from a DVS sensor mounted on a UAV.
\end{abstract}

\begin{IEEEkeywords}
Differential Evolution, Bayesian Optimisation, Self-adaptation, STDP, Neuromorphic Engineering
\end{IEEEkeywords}

\IEEEpeerreviewmaketitle

\section{Introduction}

State-of-the-art robotic unmanned aerial vehicle (UAV) systems are achieving impressive results for compact and agile flight and manoeuvring~\cite{specialissue}. However these systems are typically less power efficient and robust than their natural counterparts (e.g., bees are capable of robust flight, obstacle avoidance, and cognitive capabilities with a neural processing technology that consumes approximately 10\,$\mu$W of power, and that occupies a volume of less than 1\,mm$^3$). Using nature as inspiration, neuromorphic engineers have attempted to bridge the power-consumption gap through hardware solutions~\cite{Liu2010}. Recently, a range of different neuromorphic processors has been proposed to allow for the hardware implementation of spiking neural networks (SNNs)~\cite{chicca2014neuromorphic,indiveri2015neuromorphic,Davies_etal18,Moradi_etal17,Merolla_etal14a,Furber_etal14}. These mixed-signal analog/digital chips are ultra low power (on the order of mW) and provide an attractive alternative to current digital hardware used in mobile applications such as robotics.

Another successful neuromorphic example is the recent development of silicon retinas and event-based sensors such as the Dynamic Vision Sensor (DVS)~\cite{Lichtsteiner_etal08,Serrano-Gotarredona2013}. The DVS operates differently compared to conventional video cameras: instead of integrating light in a pixel array for a period of time and then converting it to an image, it detects local changes in luminance at each pixel and transmits these change events asynchronously as they are detected and with microsecond latency~\cite{delbruck2008frame}. Compared to standard frame-based cameras, a DVS offers (i) faster response times through asynchronous transmission, (ii) much lower bandwidth, and (iii) no motion blur~\cite{Mueggler2014}.

A system with sensors and image processing in-situ on the UAV is an essential step for autonomous UAV systems.  Typically, high-speed agile manoeuvres such as juggling, pole acrobatics, or flying through thrown hoops, use external motion sensors and high powered CPUs to control the UAVs~\cite{Muller2011,brescianini2013quadrocopter,mellinger2011minimum}; a combination of DVS and Neuromorphic spiking networks provides low-power and high response rates together with the potential for adaptation from the SNNs, and as such are promising technologies for autonomous UAVs. 

A model that has shown promise for fast and robust collision avoidance in UAV robotics applications is the locust Lobula Giant Motion Detector (LGMD)~\cite{BlanchardEtAl2000,YueRind2005,YueEtAl2006,YueRind2006,HuEtAl2017,Hartbauer2017}. The LGMD in a locust is capable of responding to an object looming at speeds ranging from $0.3$\,m$/$s to $10$\,m$/$s~\cite{rind2016two}; our model was tested on stimuli that loomed at rates of 266 pixels per second to 1478 pixels per second.  The locust uses the LGMD to escape from predators by detecting whether a stimulus is looming (increasing in size in the field of view) or not~\cite{Santer2004}. This neuronal looming detection mechanism is robust to translation, which is why it is an ideal candidate for obstacle avoidance. Previous implementations of this model used frame based cameras and simplified neural models for embedded robotic applications~\cite{Santer2004,yue2010reactive,stafford2007bio}.
Salt et al. \cite{Salt2017} modified the LGMD model from \cite{BlanchardEtAl2000} first mentioned in~\cite{rind1996neural} to use Adaptive Exponential Integrate and Fire (AEIF) neuron equations~\cite{brette2005adaptive} which model faithfully the behavior of silicon neurons present in hardware neuromorphic processors\cite{chicca2014neuromorphic}. The LGMD Neural Network (LGMDNN) was, in particular, modified to make it compatible with the Reconfigurable On-Line Learning Spiking (ROLLS) neuromorphic processor~\cite{Qiao2015}. In this previous work we have presented a proof of concept demonstration that a LGMDNN network can be used for obstacles avoidance on a UAV\@. Coupling the LGMDNN with the AEIF neural equations yields 11 user-defined parameters after making simplifying assumptions based on the constraints of the neuromorphic processor. Optimising this parameter space is challenging as it contains complex inter-dependencies. \revised{params}{Moreover, these parameters are shared by neurons and synapses in different parts of the structured neuronal network and thus have influence on the overall performance of the network that does not lend itself to any simple description of the role of each parameter.}{} Here we demonstrate the method by which the appropriate parameters were found in~\cite{Salt2017} and show the extension of this work to incorporate synaptic plasticity and neural spike-frequency adaptation mechanisms.

Identifying acceptable parameter sets for robust functional operation of this model is the focus of this work.  Due to the computational resources and time requirements involved (approximately 30 seconds to 4 minutes per evaluation), a brute force exhaustive search is unfeasible. \revised{unfeasible}{Using granularity of 1 for parameters with an upper bound greater than 100, 0.1 for an upper bound greater than 2 but less than 100, 0.01 otherwise, and the optimistic estimate of 30 seconds for each evaluation on an eight core computer yields an expected $1.87\times 10^{20}$ years to evaluate.}{}  We  set the goal to investigate the use of efficient stochastic optimisation algorithms. Differential Evolution (DE)~\cite{storn1997differential} is particularly suited to our application. DE is a simple and efficient stochastic vector-based real-value parameter optimisation algorithm with performance (at least) comparable to other leading optimisation algorithms~\cite{de-vs-pso,de-ga-compare}. DE has only two user-defined rates~\cite{das2011differential,storn1997differential,pedersen2010good}, however their optimal values are problem specific and can drastically affect algorithmic performance~\cite{brest2006self}.  This has prompted research into Self-Adaptation (SA), which allows the rates to vary autonomously in a context-sensitive manner throughout an optimisation run. Self-Adaptive DE (SADE) has been shown to perform at least as well as DE on benchmarking problems~\cite{brest2006self,qin2009differential}.  Importantly, SA has been shown to reduce the number of evaluations required per optimisation in resource-constrained scenarios with protracted evaluation times~\cite{howard2017gecco}, compared to non-adaptive solutions.  Additionally, we implemented Bayesian optimisation (BO) which models the optimisation space as a Gaussian process and uses a utility function to determine which points to select for optimisation\cite{brochu2010tutorial}. We compare these optimisers to a uniform random search as a baseline as it has been shown to outperform grid search in some problems~\cite{bergstra2012random}.

Spiking networks are particularly amenable to a form of unsupervised learning called Spike-Timing Dependent Plasticity (STDP)~\cite{bi-poo}, which allows synaptic weights to change autonomously in response to environmental inputs.  STDP has been shown to provide faster responses compared to non-plastic networks in dynamic environments~\cite{howard2012evolution}, which motivates our investigations into its use in our LGMD networks.

The combination of Evolutionary optimisation with learning, e.g., the Baldwin Effect, is known to be beneficial in artificial and natural systems~\cite{baldwin}.  In our case off-line optimization (DE or BO) sets up network parametrizations for STDP (learning) to exploit on-line.  

Specific use of STDP with meta-optimization is a promising and entirely new area for LGMD networks, which motivates our work in this area.  Our hypothesis is that these adaptivity mechanisms are beneficial to the performance of the LGMD network.  To test this hypothesis, we evaluate the performance of our optimizers (DE, SADE, and BO, with and without STDP) when optimising looming responses in LGMD networks which are stimulated by (i) simple and (ii) complex DVS recordings on the UAV. Our finding is, however, that the STDP can be both beneficial and destructive for performance of the network.

The original contributions of this work are (i) the development of an objective function that accurately describes the desired LGMD behaviour, (ii) a comprehensive statistical comparisons of three leading algorithms in optimising LGMD networks,  and (iii) the first optimisation-based study on the effect of STDP and spike-frequency adaptation in LGMD networks.

\section{The Model}
This section describes the background for the model set-up, and the specific equations that were used in the experiment. 
\subsection{LGMD}
\begin{figure}
\centering
\includegraphics[width=0.75\columnwidth]{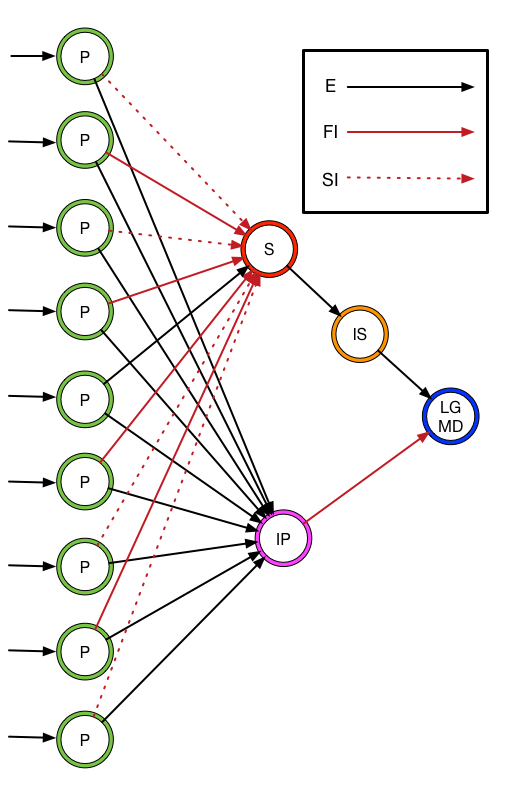}
\caption{The neuromorphic LGMD model, which consists of a photoreceptor layer (P), a summing layer (S), an intermediate photoreceptor layer (IP), an intermediate summing layer (IS), and an LGMD neuron layer.  Edges show connections that are either Excitatory (E), Slow Inhibitory (SI), or Fast Inhibitory (FI).}
\label{fig:LGMDMe}
\end{figure}

We implement the model as described by Salt et al.~\cite{Salt2017}. The LGMD model consists of a photoreceptor layer (P), a summing layer (S), an intermediate photoreceptor layer (IP), an intermediate summing layer (IS), and an LGMD neuron layer. The intermediate layers can be seen as analogous to sum-pooling layers in deep convolutional neural networks~\cite{liu2015treasure,babenko2015aggregating,fernando2017rank}. These layers are modelled as populations of AEIF neurons,  connected by  excitatory (E), slow inhibitory (SI), and fast inhibitory (FI) connections. \fig{fig:LGMDMe} shows the topology of the network.

\revised{R1_2}{The P to IP to LGMD connection inhibits the spiking response of the LGMD to translational motion across the field of vision, and the inhibitory connections (SI and FI) from the photoreceptor to the summing layer inhibit the output neuron from spiking in response to non-looming stimuli.}{} The weights of the inhibitory connections are assigned based on their distance from the central excitatory neuron similarly to that described in~\cite{BlanchardEtAl2000}. This connection configuration spans the P layer like a kernel.

The intermediate layers were added to make the model compatible with the dynap-se neuromorphic processor described in~\cite{Moradi_etal17,indiveri2015neuromorphic}. However, Salt et al.~\cite{Salt2017} found that the addition of the intermediate (sum-pooling) layer before the LGMD neuron also increased the performance of the network on all but slow circular stimuli. 

\subsubsection{Adaptive Exponential Integrate and Fire Spiking Networks}
We use Adaptive Exponential Integrate and Fire (AEIF) model neurons in the network; the respective neuron equations follow (\ref{eq:gerstnerAEIF}) and (\ref{eq:current}):

\begin{equation}
\label{eq:gerstnerAEIF}
\frac{dV}{dt} = \frac{-g_L(V-E_L)+g_L\Delta_T\exp(\frac{V-V_T}{\Delta_T})+I}{C},
\end{equation}

\begin{equation}
\label{eq:current}
I=I_e-I_{iA}-I_{iB}-I_{adapt},
\end{equation}
where $C$ is the membrane capacitance, $g_L$ is the leak conductance, $E_L$ is the leak reversal potential, $V_T$ is the spike threshold, $\Delta_T$ is the slope factor, $V$ is the membrane potential, $I_{e}$ is an excitatory current, $I_{adapt}$ is the adaptation current, and $I_{iA}$ and $I_{iB}$ describe fast and slow inhibitory current, respectively~\cite{brette2005adaptive}. When a spike is detected ($V>V_T$) the voltage resets ($V=V_r$), and the post-synaptic neuron receives a current injection from the pre-neuron firing given by:
\begin{align}
\label{eq:inject}
I_{e/i,l} &= I_{e/i,l} + q_{e/i,l},\\
I_{adapt} &=I_{adapt}  + b,
\end{align}
where the subscript $l$ corresponds to the post-synaptic layer, $q_{e/i,l}$ is the injected current, $b$ is the spike-triggered contribution to adaptation, and  the subscript $e/i$ refers to either excitation or inhibition. To simplify the model for embedded implementation, inhibitory currents were set as a ratio of the excitatory current:
\begin{equation}
q_{i,l} = inh_l\cdot q_{e,l},
\end{equation}
where $inh_l$ is a constant parameter. This equation holds for both types of inhibition (slow and fast). The decay of the excitatory or inhibitory currents is described by:

\begin{equation}
\frac{dI_{e/i}}{dt}=-\frac{I_{e/i}}{\tau_{e/i}},\label{eq:currentDecay}
\end{equation}
where $I_{e/i}$ is the current and $\tau_{e/i}$ is the time constant for the decay. The subscript $e/i$ refers to either excitation or inhibition respectively. Finally, the decay of the adaptation current is described by:
\begin{equation}
\frac{dI_{adapt}}{dt}=\frac{a(V-E_L)-I_{adapt}}{\tau_{adapt}},
\end{equation}
where $a$ is the sub-threshold adaptation and $\tau_{adapt}$ is the time constant for the decay.

\secn{ssec:expsu} explains how these were implemented and the bounds of all of the values. 

\subsection{Spike Timing Dependent Plasticity}
Spike Timing Dependent Plasticity (STDP) is a realisation of Hebbian learning based on the temporal correlations between pre- and post-synaptic spikes. This synaptic plasticity is thought to be fundamental to adaptation, learning, and information storage in the brain~\cite{song2000competitive,sjostrom2010spike}.

\begin{figure}[htbp]
\centering
\includegraphics[width=0.8\columnwidth]{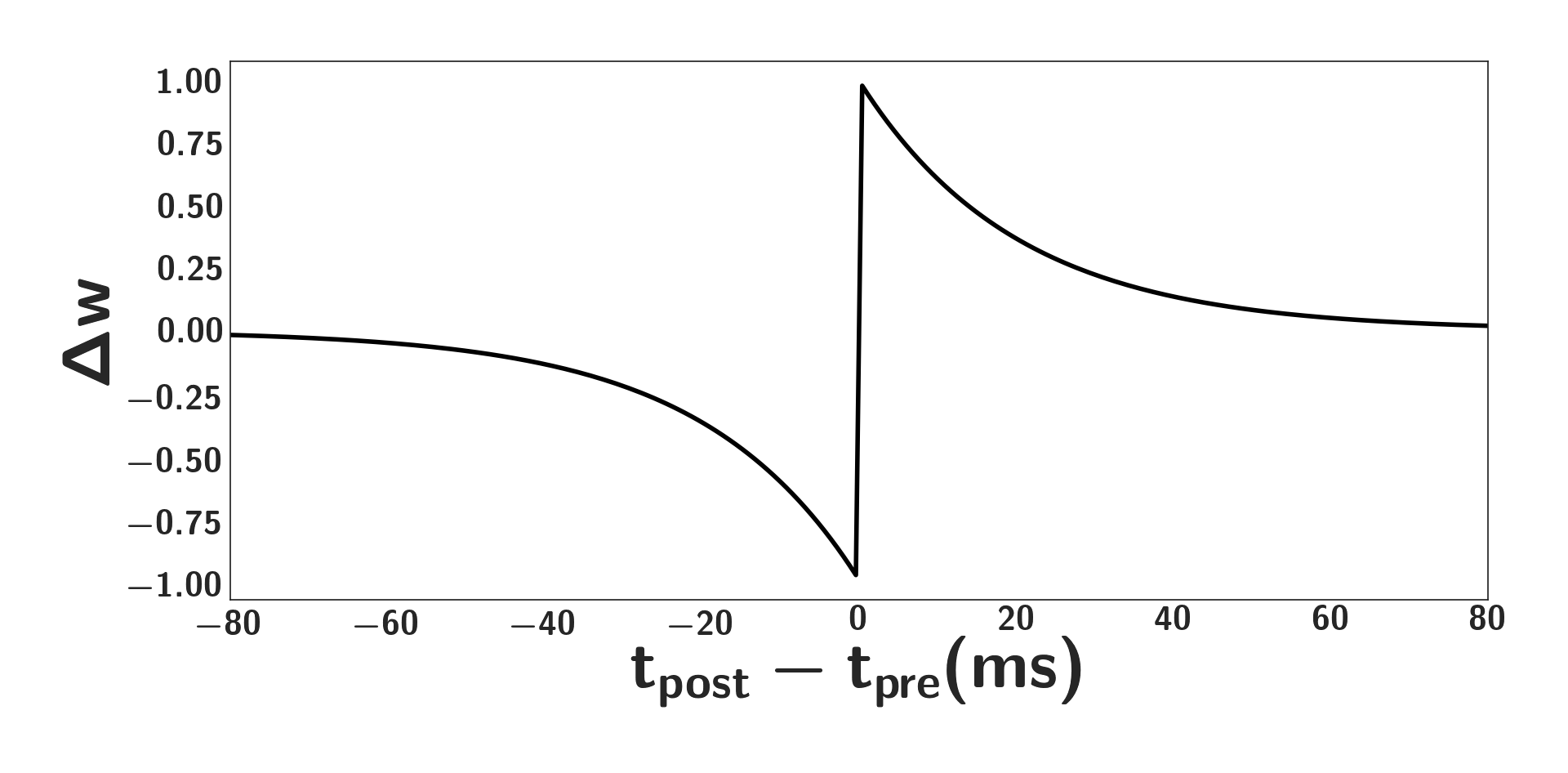}
\caption{The impact of STDP on the synaptic weights. If the pre-synaptic spike arrives before the post synaptic spike, then the strength of the weights is increased. If the post synaptic spike arrives first, then the strength of the synapse is weakened.}\label{fig:STDP}
\end{figure}

Arrival of a pre-synaptic spike closely before a post-synaptic spike increases the efficacy of the synapse, while  if a post-synaptic spike is received in close proximity to and before a pre-synaptic spike, the efficacy of the synapse is decreased.  Long term potentiating (LTP, synaptic weight increase) of the synapse occurs in the former case, long term depression (LTD, synaptic weight decrease) occurs in the latter case. \fig{fig:STDP} shows a typical dependence of the synaptic weight change on the difference in arrival times of the post- and pre- synaptic spikes. 

STDP modifies the synaptic current injection given in (\ref{eq:inject}) by multiplying it by a weight $w$, which is modified according to the plasticity rule. In particular, if a pre-synaptic spike occurs, then:
\begin{align}
I_{e/i,l} &= I_{e/i,l}+ w q_{e/i,l},\\
A_{pre} &= A_{pre}+ \Delta_{pre,}\\
w &= w + A_{pre}.
\end{align}
If a post-synaptic spike occurs then:
\begin{align}
A_{post} &= A_{post} + \Delta_{post},\\
w &= w - A_{post}.
\end{align}
$A_{pre|post}$ are the amount by which the weight $w$ is strengthened or weakened, and $\Delta_{pre|post}$ is a user-defined value for increasing $A_{pre|post}$ each time a spike occurs.
At each spike event, the variables $A_{pre|post}$ decay:
\begin{align}
\frac{dA_{pre|post}}{dt} &= -\frac{A_{pre|post}}{\tau_{pre|post}}.
\end{align}
This learning rule leads to potentiation of synapses that are supported by temporal sequence of pre- and post-synaptic spikes and depression of synapses that connect neurons that fire in a reverse order. In other words, the connection strengths vary depending on the activity of the neurons they are connected to.

\section{Optimisation Techniques}

In this Section, we describe the three optimisation techniques that we compare: DE, SADE, and BO, and how they are applied to optimising the LGMDNN parameter space. \revised{params3}{Each individual, $x_i$, is a parametrisation of the LGMDNN, given by:
$x_i =$[$\mathbf{\tau_e }$, $\mathbf{\tau_{iA} }$, $\mathbf{\tau_{iB} }$, $\mathbf{q_{eP} }$, $\mathbf{q_{eS} }$, $\mathbf{q_{eIP} }$, $\mathbf{q_{eIS} }$, $\mathbf{q_{eL} }$, $\mathbf{inhA_S}$, $\mathbf{inhB_S}$, $\mathbf{inhA_L}$, [$\mathbf{a}$, $\mathbf{b}$, $\mathbf{\tau_{w_{adapt}} }$], ($\mathbf{\tau_{pre} }$, $\mathbf{\tau_{post }}$, $\mathbf{\Delta_{pre}}$, $\mathbf{\Delta_{post}}$)]. The bounds for each element of $x_i$ can be found in \tab{tab:Bounds} in Section~\ref{ssec:expsu}, as well as a brief explanation of the meaning of each parameter.}{}

\subsection{Differential Evolution}
DE is an efficient and high performing optimiser for real-valued parameters~\cite{das2011differential,storn1997differential}. As it is based on evolutionary computing, it performs well on multi-modal, discontinuous optimisation landscapes.  Storn and Price~\cite{storn1997differential} showed that their original DE outperformed several other stochastic optimization techniques in benchmarking tests whilst requiring the setting of only two parameters, crossover probability $CR$ and differential weight $F$. It also requires a mutation function, which determines how individuals in the population are mixed. Many variants of the mutation function have been suggested, these follow the naming convention $DE/x/y/z$.  Here we use DE rand/1/bin where $x$ denotes the vector to be mutated (in this case a random vector), $y$ denotes the number of vectors used, and $z$ denotes the crossover method (bin corresponds to binomial).

The initial population, $\mathbf{X_1}=\{\mathbf{x}_{1,1}, \mathbf{x}_{2,1},...,\mathbf{x}_{NP,1}\}$, where $NP$ is the size of the population and $\mathbf{x}_{i,1}\in \mathcal{R}^D$ is an individual that contains the $D$ parameters to be optimised, is generated from random samples drawn from a uniform probability distribution of the parameter space, bounded to the range of the respective variable. These bounds are shown in \secn{sssec:hpc}. The fitness of each vector in the population is calculated by the objective function, as described in \secn{sec:OF}.

In each generation, each parent generates one offspring by way of a `donor' vector, created following Eq.~(\ref{eq:DErand1bin}):

\begin{equation}
\label{eq:DErand1bin}
v_{i,G+1}=x_{r_1,G}+F\cdot (x_{r_2,G}-x_{r_3,G}),
\end{equation}
where $r_1\neq r_2\neq r_3\neq i\in[1,NP]$ index random unique population members, the subscript $G$ indicates the current generation, and differential weight $F\in[0,2]$ determines the magnitude of the mutation. The final offspring is generated by probabilistically merging elements of the parent with elements of the donor vector. The new vector $u_{i,G+1} = (u_{1i,G+1},\ldots , u_{Di,G+1})$ is found by:

\begin{equation}
u_{ji,G+1}=\begin{cases}
v_{ji,G+1},& \text{if }rand(j)\leq CR \text{ or } j = R,\\
x_{ji,G},& \text{otherwise},
\end{cases}
\end{equation}
where $j \in (1,\ldots,D)$, $CR\in [0,1]$ is the crossover rate, $rand(j)\in [0,1]$ is a uniform random number generator, and $R\in (1,\ldots,D)$ is a randomly chosen index to ensure that at least one parameter changes.  The value of offspring with index $i$ is then calculated as:

\begin{equation}
x_{i,G+1} = \begin{cases}
u_{i,G+1},& \text{if }f(u_{ji,G+1})>f(x_{i,G}),\\
x_{i,G},&\text{otherwise},
\end{cases}
\end{equation}

\revised{fitness_def}{where $f(\cdot)$ is the fitness function.}{} Once all offspring are generated, they are evaluated on the fitness function, and selected into the next generation if they score better than their parent.  Otherwise, the parent remains in the population.

\subsection{Self-Adaptive DE}
\label{sec:SADE}

Brest et al.~\cite{brest2006self} present the first widely-used self-adaptive rate-varying DE, which is expanded by Qin et al., to allow the mutation scheme to be selected (from four predetermined schemes) alongside the rates~\cite{qin2009differential}, based on previously successful settings.  Different rates/schemes are shown to work better on different problems, or in different stages of a single optimisation run.  The strategy for a given candidate is selected based on a probability distribution determined by the success rate of a given strategy over a learning period $LP$. A strategy is considered successful when it improves the individual's value.  In the interest of brevity, we refer the interested reader to~\cite{qin2009differential} for a full algorithmic description.  

\revised{plot_drop}{Rates are adapted as follows. Before $G>LP$ (where $G$ is number of generations, and $LP$ is the number of generations needed before the learned $CR$ values are used), CR is calculated by randomly selecting a number from a normal distribution, $N(0.5,0.3)$, with a mean of 0.5 and a standard deviation of 0.3. Afterwards it is calculated by a random number from $N(CR_{mk},0.1)$ where $CR_{mk}$ is the median value of the successful $CR$ values for each strategy $K$. $F$ is simply selected from a normal distribution $N(0.5,0.3)$, which will cause it fall on the interval $[-0.4,1.4]$ with a probability of 0.997~\cite{qin2009differential}.}{}

\subsection{Bayesian Optimisation}

Bayesian optimisation (BO), e.g.~\cite{brochu2010tutorial}, is a probabilistic optimisation process that typically requires relatively few evaluations~\cite{mockus1994application, jones2001taxonomy, sasena2002flexibility}, although the evaluations themselves are computationally expensive. When parallelised, BO is shown to locate hyper-parameters within set error bounds significantly faster than other state-of-the-art methods on four challenging ML problems\cite{snoek2012practical}, in one case displaying 3\% improved performance over state-of-the-art expert results. As such, BO can be considered a competitive optimiser to which we can compare DE and SADE.

BO assumes the network hyper-parameters are sampled from a Gaussian process (GP), and updates a prior distribution of the parameterisation based on observations. For LGMDNN, observations are the measure of generalization performance under different settings of the hyper-parameters we wish to optimise.  BO exploits the prior model to decide the next set of hyper-parameters to sample. 

BO comprises three parts: (i) a prior distribution, (ii) an acquisition function, and (iii) a covariance function.

\subsubsection{Prior}
We use a Gaussian Process (GP) prior, as it is particularly suited to optimisation tasks~\cite{mockus1994application}.  A GP is a distribution over functions specified by its mean, $m$, and covariance, $k$, which are updated as hyper-parameter sets are evaluated.  The GP returns $m$ and $k$ in place of the standard function $f$:

\begin{equation}
f(x)\sim GP(m(x),k(x,x')).
\end{equation}

\subsubsection{Covariance Function}
The covariance function determines the distribution of samples drawn from the GP~\cite{brochu2010tutorial,snoek2012practical}.  Following \cite{snoek2012practical}, we select the 5/2 ARD Mat{\' e}rn kernel (\ref{eq:matern}), where $\theta$ is the covariance amplitude.

\begin{equation}
\label{eq:matern}
k_{m52}(x_i,x_j) = \eta exp(-\sqrt{5r^2(x_i,x_j)}),
\end{equation}
where:
\begin{equation}
\eta=\theta(1+\sqrt{5r^2(x_i,x_j)}+\frac{5}{3}r^2(x_i,x_j)),
\end{equation}
where:
\begin{equation}
r^2(x_i,x_j)=\frac{x_i-x_j}{\theta^2}.
\end{equation}

\subsubsection{Acquisition Function}
An acquisition function is a function that selects which point in the optimisation space to evaluate next. We evaluate the three acquisition functions, which select the hyper-parameters for the next experiment: Probability of Improvement (PI), Expected Improvement (EI)\cite{mockus1994application}, and Upper Confidence Bound (UCB)\cite{srinivas2009gaussian} --- see ~\cite{brochu2010tutorial} for full implementation details. $\mu(\cdot)$ and $\sigma(\cdot)$ refer to the mean and standard deviation functions. 

Briefly, the {\bf PI} can be calculated, given our current maximum observation of the GP, $x^+$, by:

\begin{align}
PI(x) &= P(f(x)\geq f(x^+)+\zeta )\nonumber \\ &= \Phi(\frac{\mu(x)-f(x^+)-\zeta}{\sigma(x)}).\label{eq:PIZ}
\end{align}
Where, $\Phi (.)$ is the normal cumulative distribution function.
Here, $\zeta\geq 0$ is a user-defined trade-off parameter that balances exploration and exploitation~\cite{lizotte2008practical}.

Similarly, {\bf EI} is evaluated by ~\cite{jones1998efficient}:

\begin{align}
EI(x) &= \begin{cases}
ei+\sigma(x)\phi(Z), &\text{if } \sigma(x)>0,\\
0, & \text{otherwise};
\end{cases}\label{eq:EI}\\
ei&=(\mu(x)-f(x^+)-\zeta)\Phi(Z);\\\nonumber
Z &= \begin{cases}
\frac{\mu-f(x^+)-\zeta}{\sigma(x)},&\text{if } \sigma(x)>0,\\
0, & \text{otherwise},
\end{cases} \nonumber
\end{align}
where $\phi$ and $\Phi$ correspond to the probability and cumulative distribution functions of the normal distribution, respectively. 

{\bf UCB} maximises the upper confidence bound:
\begin{equation}
UCB(x) = \mu(x) + \kappa \sigma(x),
\end{equation}
where $\kappa \geq 0$ balances exploration and exploitation~\cite{snoek2012practical}, and is calculated per evaluation as: 
\begin{equation}
\kappa = \sqrt{\mathit{\nu \tau_t}},
\end{equation} 
where $\nu$ is the user tunable variable and:
\begin{equation}
\tau_t = 2\log(\frac{t^{\frac{d}{2}+2}\pi^2}{3\delta}).
\end{equation} 
$\delta \in \{0,1\}$, $d$ is the number of dimensions in the function and $t$ is the iteration number.

\section{Test Problem}
This section outlines the rationale of the objective function, the experimental set-up, and assumptions. It is important to note that the motivation behind the model simplifications and objective function is for the work to be directly transferable to the neuromorphic processors described in~\cite{Qiao2015} once they are readily available.

\subsection{Objective Function}
\label{sec:OF}

Initially the optimisation function was formulated as:
\begin{equation}
F_{init}(\lambda) = Acc-||V||_2 ,
\end{equation}
where $Acc$ is the accuracy given by \eq{eq::acc}, $||V||_2$ is \revised{fitness2}{the regularisation term, in particular the  l2-norm of the voltage signal, used to regularise the voltage signal, and $\lambda$ is a candidate solution.}{} However, this objective function resulted in all optimisers producing 50\% accuracy with looming detected at any time in the experiments.

To improve the accuracy, the function to optimise was formulated as a weighted multi-objective function~\cite{deb2001multi}. The objective function has three distinct parts: accuracy (\eq{eq::acc}), sum squared error of the membrane voltage signal (\eq{eq::SSEOS}), and the reward of the spiking trace (\eq{eq::reward}). The accuracy alone could not be used as there were only eight discrete events in the input stimuli during the optimisation phase which was not granular enough for optimisation. \eq{eq::SSEOS} acts as a regularising term to prevent the voltage trace from becoming too large. \eq{eq::reward} is used to rate the spiking behaviour with more granularity than is possible with accuracy. Combining \eq{eq::SSEOS} and \eqref{eq::reward} resulted in spiking behaviour with a realistic voltage trace. The accuracy was then used to account for sub-optimal regions of \eqref{eq::reward} that still resulted in a high score. This resulted in the final formulation of the objective function modified by accuracy, $F_{Acc}(\lambda)$, which is calculated by:
\begin{equation}
\label{eq::Facc}
F_{Acc}(\lambda) = \begin{cases}
2\cdot F(\lambda), & \text{if } F(\lambda)>0\text{ and } Acc=1,\\
Acc\cdot F(\lambda), & \text{if } F(\lambda)<0,\\
0, & \text{if } Acc=1\text{ and }F(\lambda)<0,\\
F(\lambda), & \text{otherwise}.
\end{cases}
\end{equation}
Here, $Acc$ is the accuracy of the LGMDNN output and $F(\lambda)$ is the fitness function. The LGMD network is said to have detected a looming stimulus if the output neuron's spike rate exceeds a threshold $SL$. This can be formalised by:
\begin{equation}
Looming = \begin{cases}
\text{True}, &\text{if } SR>SL,\\
\text{False}, &\text{otherwise},
\end{cases}
\end{equation}
where $SR$ can be calculated by:
\begin{equation}
\label{eq::spikeRate}
SR = \sum^{t+\Delta T}_{i=t} S_i,
\end{equation} 
where $\Delta T$ is the time over which the rate is calculated and $S_i$ is whether or not there is a spike at time $i$; a spike is defined to occur if at time $i$ the membrane potential exceeds $V_T$ ($V_T$ has the same meaning as in \eq{eq:gerstnerAEIF}).

The looming outputs are  categorised into true positives ($TP$), false positives ($FP$), true negatives ($TN$), and false negatives ($FN$). Output accuracy is then:
\begin{equation}
\label{eq::acc}
Acc = \frac{TP+TN}{TP+TN+FP+FN}.
\end{equation}

The fitness function without accuracy, $F(\lambda)$, can be calculated by:
\begin{equation}
F(\lambda) = \frac{Score(\lambda)+SSEOS(\lambda)}{2}, 
\end{equation}
where {\em Score} is a scoring function based on the timing of spiking outputs and $SSEOS$ is the sum squared error of the output signal. 

The score is calculated by difference of the penalties' and reward functions' sums over the simulation:
\begin{equation}
Score(\lambda) = \sum^N_{i=1}R_i - \sum^N_{i=1}P_i.
\end{equation}

The reward at a given time can be calculated by:
\begin{equation}
\label{eq::reward}
R(t) = \begin{cases}k\exp(\frac{t}{\Delta t})+1, & \text{if looming and spike},\\
0, & \text{otherwise}.
\end{cases}
\end{equation}

The punishment can be calculated by:
\begin{equation}
P(t) =
\begin{cases}
(l-c)\frac{t}{\Delta t}+c,& \text{if not looming and} \\
\ & \text{spike and } t<\frac{\Delta t}{2};\\
(l-c)\frac{1-(t-\frac{\Delta t}{2})}{\frac{\Delta t}{2}}+c, & \text{if not looming} \\
\ & \text{and spike and } t>\frac{\Delta t}{2};\\
0, & \text{otherwise}.
\end{cases}
\end{equation}
In these equations, $t$ and $\Delta t$ remain consistent with the other objective functions and $k$, $l$, and $c$ are all adjustable constants to change the level of punishment or reward.

To calculate $SSEOS(\lambda)$, the signal was first processed so that every spike had the same value. This was done so that the ideal voltage and the actual voltage would match in looming regions, as the voltage can vary for a given spike. Ultimately, the only criterion is that the voltage has crossed the spiking threshold. In the non-looming region the ideal signal was taken to be the resting potential, which was negative for the AEIF model equation. The signal error was calculated at every time step as:

\begin{equation}
\label{eq::SSEOS}
SSEOS(\lambda) = -\sum^N_{i=1}(V_{actual}^i-V_{ideal}^i)^2.
\end{equation}

$V_{actual}$ could be obtained directly from the state monitor object of the LGMD output neuron in the SNN simulator (Brian2). $N$ in this case is the length of the simulation and $i$ indicates each recorded data point at each time step of the simulation. $V_{ideal}$ was given by:

\begin{equation}
V_{ideal} = \begin{cases}
V_{spk}, & \text{if looming},\\
V_{r},& \text{otherwise},
\end{cases}
\end{equation}
where $V_{spk}$ is the normalised value given to each spike and $V_r$ is the resting potential.

\revised{voltage}{Overall, this gives an objective function that takes into account the expected spiking behaviour, whilst penalising the system for deviating from plausible voltage values and rewarding it for accurately categorising looming and non-looming stimuli. The voltage signal was kept to realistic bounds as the model was designed to target neuromorphic processors such as the ROLLS chip~\cite{indiveri2015neuromorphic}.}{}

\subsection{Experimental Set-up}
\label{ssec:expsu}
The model was set-up using the Python Brian2 spiking neural network simulator~\cite{goodman2014brian}.

\subsubsection{Data Collection}
Data was collected using a DVS in-situ mounted on a quadrotor UAV (QUAV). Two types of data were collected: simple and real world. The simple data was synthesised using PyGame to generate black shapes on a white background that increased in area in the field of view of the DVS. This included: a fast and slow circle, a fast and slow square, and a circle that loomed then translated while increasing in speed (composite). The laptop playing the stimuli was placed in front of the hovering QUAV and the stimuli were recorded. This was done to maintain any noise that might be generated by the propellers of the QUAV.   

To challenge the model, real stimuli were also recorded: a white ball on a black slope was rolled towards the DVS from 3 different directions; a cup was suspended in the air in front of the hovering QUAV and moving towards and away from the QUAV; and a hand was moved towards and away from the DVS on the hovering QUAV. These are increasing in complexity in terms of the shapes that are presented. 

Four looming and non-looming events ($\sim25$\,s) from the composite stimulus were used to optimise the model and then the optimised model was evaluated on the other stimuli. The stimuli were chosen to show that the model generated is both shape and speed invariant.

\subsubsection{Experimental Constants}
$\Delta T$ from \eq{eq::spikeRate} was set to be $10$\,ms. A loom was said to be detected if $SR$ from \eq{eq::spikeRate} exceeded 13. This had to occur before the last 10\% of the looming sequence to allow enough time for the UAV to react. The clock in Brian was set to have $0.1$\,ms granularity. This meant that the model could react after $1.3$\,ms if the loom was intense, at most it would take $10$\,ms.

\subsubsection{Hyper-parameter constraints}
\label{sssec:hpc}
The hyper-parameters were all continuous and could range from zero to infinity. There were many regions of the parameter space that were not computable even when using a cluster with 368GB of RAM. To mitigate some of the computational difficulties, Bayesian optimisation using the expected improvement utility function (BO-EI) was used over 20 eight hour runs to find feasible regions of the hyper-parameter space allowing us to constrain the optimisation space. 

$C$, $g_L$, $E_L$, $V_T$, and $\Delta_T$ are parameters of the neuron equation, not the model, and were set as constants; performance was not impacted by setting these values and appropriately optimising the other parameters~\cite{Salt2016}: $C=124.2\,pF$, $g_L=60.05\,nS$, $E_L=-73.12\,mV$, $V_T=-3.98\,mV$, and $\Delta_T=6.71\,mV$.

\tab{tab:Bounds} shows the constraints found for the rest of the hyper-parameters. 
\begin{table}[htbp]
\centering
\caption{Parameters of the optimisation space and their constraints.} \label{tab:Bounds}
\begin{tabular}{|>{\raggedright\arraybackslash}p{30pt}||>{\centering\arraybackslash}p{30pt}|>
{\centering\arraybackslash}p{30pt}|>{\centering\arraybackslash}p{100pt}||}
\hline
 \textbf{Param.}& \textbf{Min} & \textbf{Max} & \textbf{Description} \\
\hline\hline
 $\mathbf{\tau_e}$ & 1 & 10 & \small{Decay time const, exc.} \\\hline
 $\mathbf{\tau_{iA}}$ & 1 & 20 & \small{Decay time const, A inh.} \\\hline
 $\mathbf{\tau_{iB}}$ & 1 & 25 &  \small{Decay time const, B inh.}\\\hline
 $\mathbf{q_{eP}}$ & 0 & 1363 & \small{Exc. current inj. to P} \\\hline
 $\mathbf{q_{eS}}$ & 0 & 5000 & \small{Exc. current inj. to S} \\\hline
 $\mathbf{q_{eIP}}$ & 0   & 230 & \small{Exc. current inj. to IP} \\\hline
 $\mathbf{q_{eIS}}$ & 0 & 270 & \small{Exc. current inj. to IS}\\\hline
 $\mathbf{q_{eL}}$ & 0	& 472 & \small{Exc. current inj. to L}\\\hline
 $\mathbf{inhA_S}$ &0.04&1.22 & \small{Inh.A/Exc. ratio for S} \\\hline
 $\mathbf{inhB_S}$ &0.24&1.5 & \small{Inh.B/Exc. ratio for S} \\\hline
 $\mathbf{inhA_L}$ &0.019&1.3 & \small{Inh.A/Exc. ratio for L} \\\hline
 $\mathbf{a}$ &1&8 & \small{Sub-thresh. adaptation}\\\hline
 $\mathbf{b}$ &40&141 & \small{Spike-contrib. to adapt.}\\\hline
 $\mathbf{\tau_{w_{adapt}}}$ &1&150 &  \small{Time const, adapt.}\\\hline
 $\mathbf{\tau_{pre}}$ &1&25 & \small{Time const of $A_{pre}$} \\\hline
 $\mathbf{\tau_{post}}$ &1&25 & \small{Time const of $A_{post}$} \\\hline
 $\mathbf{\Delta_{pre}}$ &1e-16&0.05 & \small{Current+ at pre-syn. spike}\\\hline
 $\mathbf{\Delta_{post}}$ &1e-16&0.05 & \small{Current+ at post-syn. spike} \\\hline
\hline
\end{tabular}
\end{table}
\subsubsection{Comparing optimisers}
SADE, DE, BO with EI(BEI), BO with POI (BPOI), and BO with UCB (BUCB) were evaluated thirty times on the same input stimulus, so that they could be statistically compared using a Mann-Whitney U test. A random search (RNG) was also ran in the same manner as a benchmark for the algorithms, which has been shown to be a natural baseline with which to compare other optimisation algorithms as it can outperform grid searches in terms of results and number of calculations~\cite{bergstra2012random}. 
The input stimulus included a black circle on a white background performing a eight looming and eight non-looming events. The non-looming events contained a combination of a shrinking circle and a translating circle from left to right or right to left. The values of the user defined parameters were selected as:
\begin{itemize}
\item BEI and BPOI: $\zeta=0.01$;
\item BUCB: $\kappa=2.576$;
\item DE: $NP=\frac{10dim}{3}$, $F=0.6607$, $CR=0.9426$;
\item SADE: $LP=3$, $NP=\frac{10dim}{3}$, where dim is the number of hyper parameters;
\item RNG: Individuals were selected from a uniform distribution.  
\end{itemize}\par 
Note, that we have chosen all user-defined parameters based on values previously used in the literature.

The tests were run using the non-adaptive and non-plastic model with the bounds from \tab{tab:Bounds}. They were defined as having converged if they had not improved for $10\times NP$ evaluations. This meant ten generations for the DE algorithms and the same number of BO or RNG evaluations evaluations. The population size was two more than what is recommended by~\cite{pedersen2010good} for the DE algorithm. This size was chosen as it is relatively small and time was an issue. The short convergence meant that the SADE algorithm needed to have a short LP. The processor time was not included as a metric for this as the tests were run on three different computers so the results would not have been comparable.

\subsubsection{Comparing Models}
Once the best optimiser was found (a comparison of optimisers can be found in \ssec{ssec:optcomp}), the best performing optimiser, SADE, was used to optimise the following models:
\begin{description}
\item[\textbf{LGMD:}] Neuromorphic LGMD; 
\item[\textbf{A:}] LGMD with adaptation;
\item[\textbf{P:}] LGMD with plasticity;
\item[\textbf{AP:}] LGMD with adaptation and plasticity. 
\end{description}

The SADE variables were set to: $LP=3$ and $NP=10$. The optimisation process was run 10 times and the best optimiser from these ten runs was selected. The model was then tested on each input case for ten looming to non-looming or non-looming to looming transitions. The performance of each model is reported in \ssec{ssec:modcomp}.

Plasticity was found to degrade the performance sometimes so we experimented clamping it from 0\% to 100\% of the original synaptic strength. This allowed the weights to range from zero to double the original values when at 100\% STDP to no variation at 0\% STDP.

\section{Results and Discussion}
The results are split into two subsections. First, we will compare the optimisers and then we will compare the addition of adaptation, plasticity, and adaptation and plasticity combined to the baseline model.

The models are evaluated on their accuracy (Acc), sensitivity (Sen), Precision (Pre), and Specificity (Spe). Acc is defined in \ssec{sec:OF}. The other metrics can be found in~\cite{alpaydin}.

\subsection{Optimiser comparison and statistical analysis}
\label{ssec:optcomp}
\tab{tab:OptimiserResults} shows that the SADE algorithm achieved the best fitness. \revised{fitness3}{A good fitness value is one that is greater than 0, this means that it is either 100\% accurate or exhibits a desirable voltage trace}{} The optimisers all achieved a negative fitness values due to decreased population size and convergence conditions to enable us to run enough experiments to be able to perform statistical analysis.~\secn{ssec:modcomp} shows the results of SADE when run with a less restrictive convergence condition. The DE algorithm converged on the worst solution in the least number of objective function evaluations, and achieved the best specificity but the worst fitness, precision, and sensitivity. RNG achieved the highest average accuracy, sensitivity, and precision. BPOI and SADE still had greater fitness. As the fitness function is negative it is likely that it was rarely modified by the accuracy as per \eq{eq::Facc}.  All of the models were able to find locations in the optimisation space with 100\% accuracy/sensitivity given enough time, however as evaluations are time-expensive, not all these algorithms may  be viable and a combination of performance metrics should be considered. 

\begin{table}[htbp]
\centering
\caption{Optimisation algorithm metrics.\label{tab:OptimiserResults}}
\begin{tabular}{|l||c|c|c|c|c|c||}
\hline
\textbf{Meth} & \textbf{Fit} & \textbf{Eva} & \textbf{Acc} & \textbf{Sen} & \textbf{Pre} & \textbf{Spe} \\
\hline\hline
 \textbf{DE}	&  -3779.96 &    \textbf{440.40} &   	0.52 &          0.06 &        0.18 & \textbf{0.98} \\
 \textbf{BEI}	&  -2653.45 &           761.29 &       	0.65 &          0.39 &        0.70 &          0.90 \\
 \textbf{SADE}  &  \textbf{-1749.83} &  1345.66 &       0.61 &          0.33 &        0.58 &          0.89 \\
 \textbf{RNG}   &  -2328.15 &           754.80 & \textbf{0.67} & \textbf{0.47} &\textbf{0.75} &       0.86 \\
 \textbf{BUCB}  &  -2433.38 &           832.40 &       	0.60 &          0.43 &        0.63 &          0.77 \\
 \textbf{BPOI}  &  -1968.19 &           794.27 &       	0.63 &          0.40 &        0.71 &          0.86 \\
\hline
\end{tabular}
\end{table}

 \begin{table}[htbp]
\centering
\caption{Comparison of the statistical significance of the results. }\label{tab:statistics}
\begin{tabular}{|l|l||c|c|c|c|c|c||}
\hline
 \textbf{} & \textbf{Meth} & \textbf{Fit} &\textbf{ Eva}& \textbf{Acc} & \textbf{Sen} & \textbf{Pre} & \textbf{Spe} \\
\hline
\hline
\multirow{4}{*}{\textbf{BUCB}} 
 &DE    & +         & +               & +          & +             & +           & + \\\cline{2-8}
 &BEI   & .         & .               & +          & .             & .           & + \\\cline{2-8}
 &SADE  & +         & +               & .          & +             & .           & + \\\cline{2-8}
 &RNG   & .         & +               & +          & .             & +           & + \\\cline{2-8}
 &BPOI  & +         & .               & .          & .             & .           & + \\\cline{2-8}
\hline  
\hline
\multirow{4}{*}{\textbf{DE}} 
 &BEI   & +         & +               & +          & +             & +           & + \\\cline{2-8}
 &SADE  & +         & +               & +          & +             & +           & + \\\cline{2-8}
 &RNG   & +         & +               & +          & +             & +           & + \\\cline{2-8}
 &BUCB  & +         & +               & +          & +             & +           & + \\\cline{2-8}
 &BPOI  & +         & +               & +          & +             & +           & + \\\cline{2-8}
\hline    
\hline
\multirow{4}{*}{\textbf{BEI}} 
 &DE    & +         & +               & +          & +             & +           & +  \\\cline{2-8}
 &SADE  & +         & +               & .          & .             & .           & .  \\\cline{2-8}
 &RNG   & +         & .               & .          & +             & .           & .  \\\cline{2-8}
 &BUCB  & .         & .               & +          & .             & .           & +  \\\cline{2-8}
 &BPOI   & +         & .               & .          & .             & .           & .  \\\cline{2-8}
\hline   
\hline
\multirow{4}{*}{\textbf{SADE}}
 &DE    & +         & +               & +          & +             & +           & + \\\cline{2-8}
 &BEI   & +         & +               & .          & .             & .           & . \\\cline{2-8}
 &RNG   & +         & +               & +          & +             & +           & . \\\cline{2-8}
 &BUCB  & +         & +               & .          & +             & .           & + \\\cline{2-8}
 &BPOI  & +         & +               & .          & .             & .           & . \\\cline{2-8}
\hline  
\hline
\multirow{4}{*}{\textbf{BPOI}} 
 &DE    & +         & +               & +          & +             & +           & + \\\cline{2-8}
 &BEI   & +         & .               & .          & .             & .           & . \\\cline{2-8}
 &SADE  & +         & +               & .          & .             & .           & . \\\cline{2-8}
 &RNG   & .         & .               & .          & +             & .           & . \\\cline{2-8}
 &BUCB  & +         & .               & .          & .             & .           & + \\\cline{2-8}
\hline
\hline
\multirow{4}{*}{\textbf{RNG}}
 &DE    & +         & +               & +          & +             & +           & +  \\\cline{2-8}
 &BEI   & +         & .               & .          & +             & .           & .  \\\cline{2-8}
 &SADE  & +         & +               & +          & +             & +           & .  \\\cline{2-8}
 &BUCB  & .         & +               & +          & .             & +           & +  \\\cline{2-8}
 &BPOI  & .         & .               & .          & +             & .           & .  \\\cline{2-8}
\hline  
\hline
\end{tabular}
\end{table}

\tab{tab:statistics} shows the statistical significance of the results from \tab{tab:OptimiserResults}. The method in the comparison column is compared to each method in the subsequent column. A + indicates statistically significant values and a . indicates no statistical significance. Statistical significance was defined as $p\leq 0.05$. The Mann-Whitney U test was used to determine statistical significance because it does not require normally distributed samples.

SADE's  fitness was significantly better compared to all optimisers, but it also performed the most evaluations. This difference is statistically significant. SADE had the second best accuracy but this is only significantly different to RNG and DE. Its sensitivity was significantly worse than BUCB and RNG. It had significantly worse precision than RNG, but significantly better specificity.

BPOI was significantly better than DE, BEI, and BUCB for fitness. It also had significantly less evaluations than SADE.

BEI had significantly worse fitness and sensitivity when compared to BOPI, RNG and SADE. Interestingly, even though it had the second best accuracy the difference is only significant between it and BUCB and SADE. 

DE took significantly fewer evaluations to converge when compared to all algorithms but also had significantly worse fitness, accuracy, sensitivity, and precision. However it had significantly better specificity than all other algorithms. DE was statistically different to every algorithm for every metric.

BUCB had slightly worse fitness than RNG but it is not statistically significant. Its accuracy was significantly worse than BEI but significantly better than DE.

A possible reason that DE underperformed is that the $F$ values provided in \cite{pedersen2010good} are not appropriate for this problem. The population size may have also been too small, as populations were truncated to one third of the size recommended in~\cite{pedersen2010good}. A smaller population was used for a fair comparison to the Bayesian optimisation algorithms which have a higher computational overhead than DE and SADE. SADE outperformed DE with the same sized population and may have performed better given a larger population. Increasing the population size meant that BPOI and BUCB were not able to complete one run in the time it took SADE, DE, and RNG to do thirty. EI was only able to complete 25 of 30 runs with the larger population. The Bayesian optimisers don't have a population size but the stopping condition was based on it.  

SADE removes the need to find control parameters and has been shown to perform as well or better than DE even when the control parameters are well selected~\cite{qin2009differential}. The generalisability that comes with finding the right control parameters on-the-fly is also appealing.

The addition of the various mutation functions to SADE also seems to help it find better results. This is likely due to the desirable properties of each mutation function cancelling out the undesirable properties of other mutation functions.

A surprising result was that of the BO algorithms BPOI seemed to perform the best. This contrasts to previous studies~\cite{brochu2010tutorial} that ranked it last compared to BEI and BUCB. BPOI tends to focus more on exploitation rather than exploration, choosing regions in the GP with a higher mean rather than variance.

\fig{fig:SADEALL} also shows that SADE's CR and F values converged on small values indicating that it also preferred exploitation to exploration. We postulate that the surface has multiple sharp peaks making it difficult for the optimisers to find good values, but finding one of these peaks and climbing it yields better results than being overly exploratory. For example, if BO landed near a sharp peak using UCB or EI it would never be inclined to evaluate nearby points, because the utility function tends to prioritise regions of higher uncertainty and uncertainty may be lower near evaluated points. RNG offers an unbiased search through the space and performs worse, in terms of fitness, than the exploitative algorithms but better than the exploratory ones. 

\subsection{SADE Averages}
\begin{figure}[h!tbp]
  \centering
  \begin{subfigure}{0.9\columnwidth}
    \centering
    \includegraphics[width=1\columnwidth]{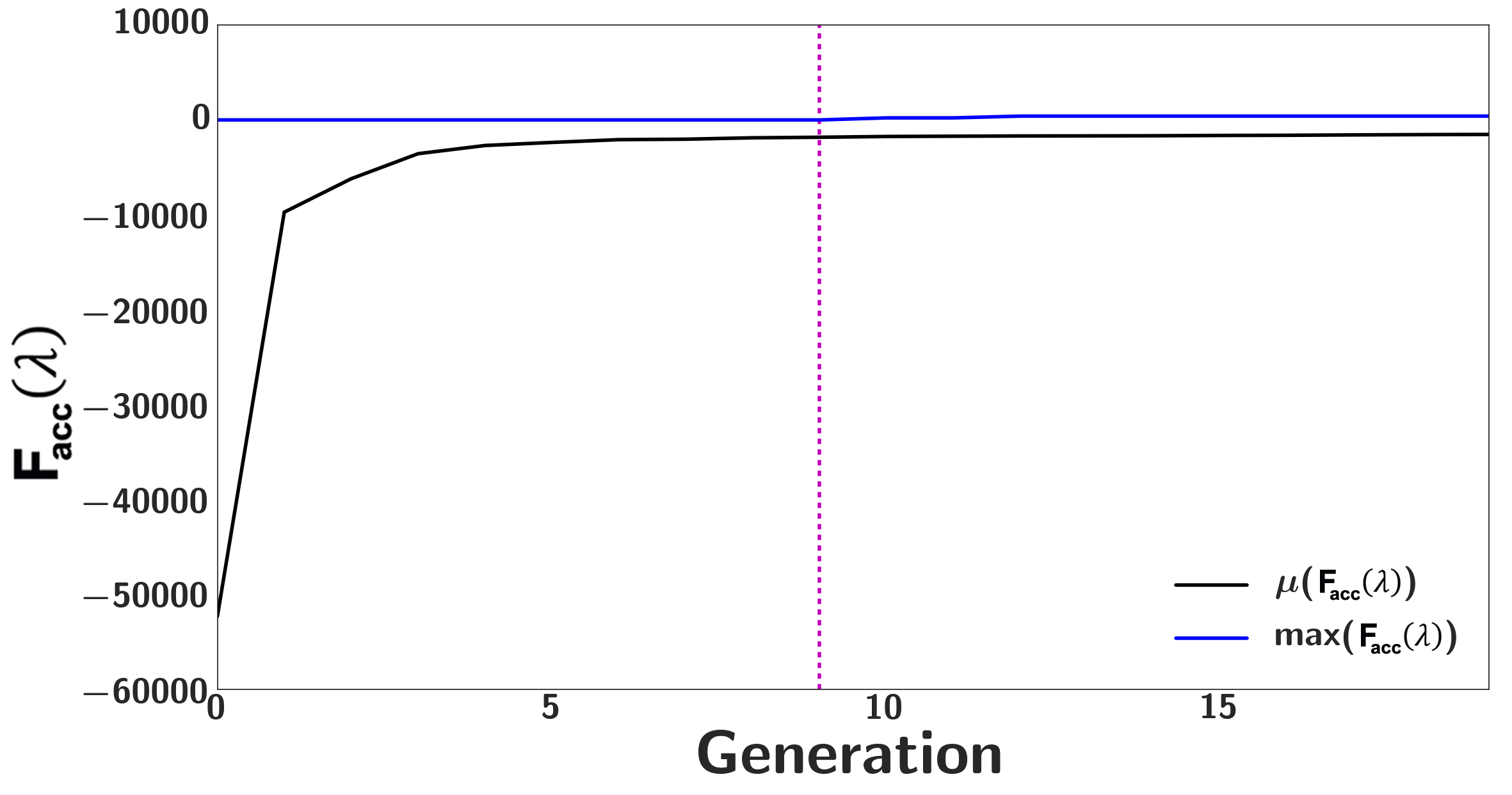}
    \caption{}\label{fig:sAVG}
  \end{subfigure}
  \begin{subfigure}{0.9\columnwidth}
    \centering
    \includegraphics[width=1\columnwidth]{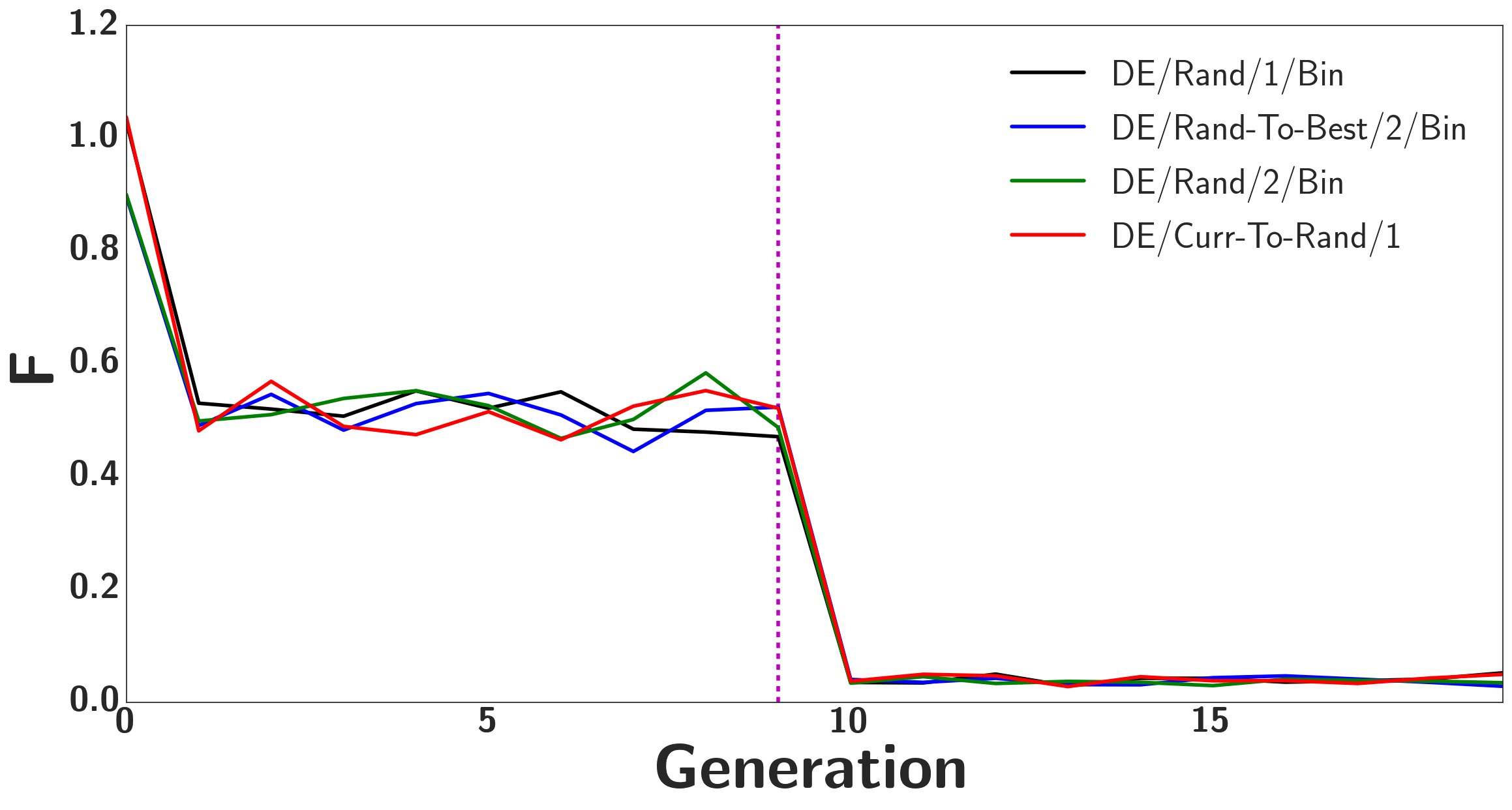}
    \caption{}\label{fig:fAVG}
  \end{subfigure}
  \begin{subfigure}{0.9\columnwidth}
    \centering
    \includegraphics[width=1\columnwidth]{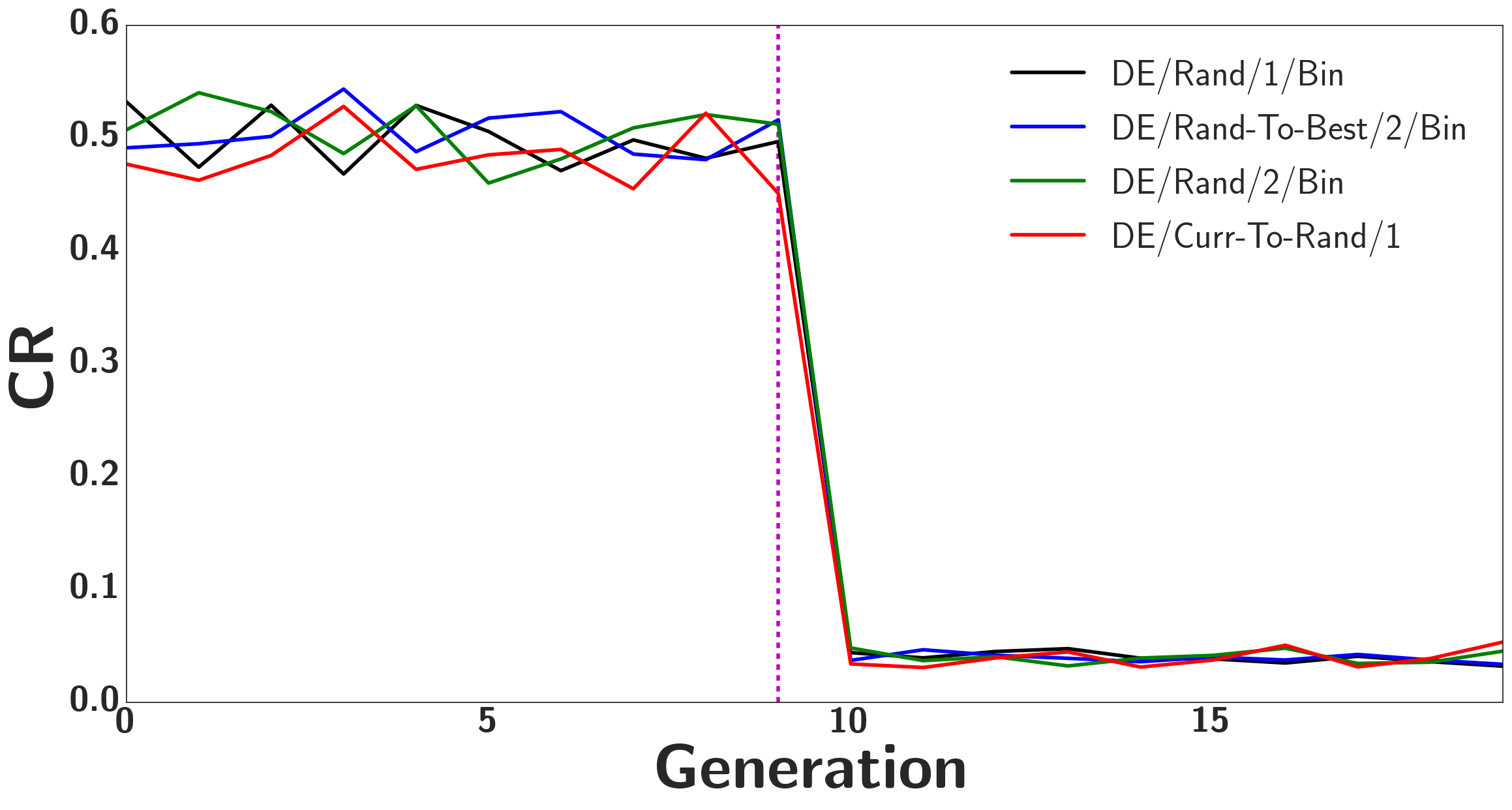}
    \caption{}\label{fig:crAVG}
  \end{subfigure}
  \begin{subfigure}{0.9\columnwidth}
    \centering
    \includegraphics[width=1\columnwidth]{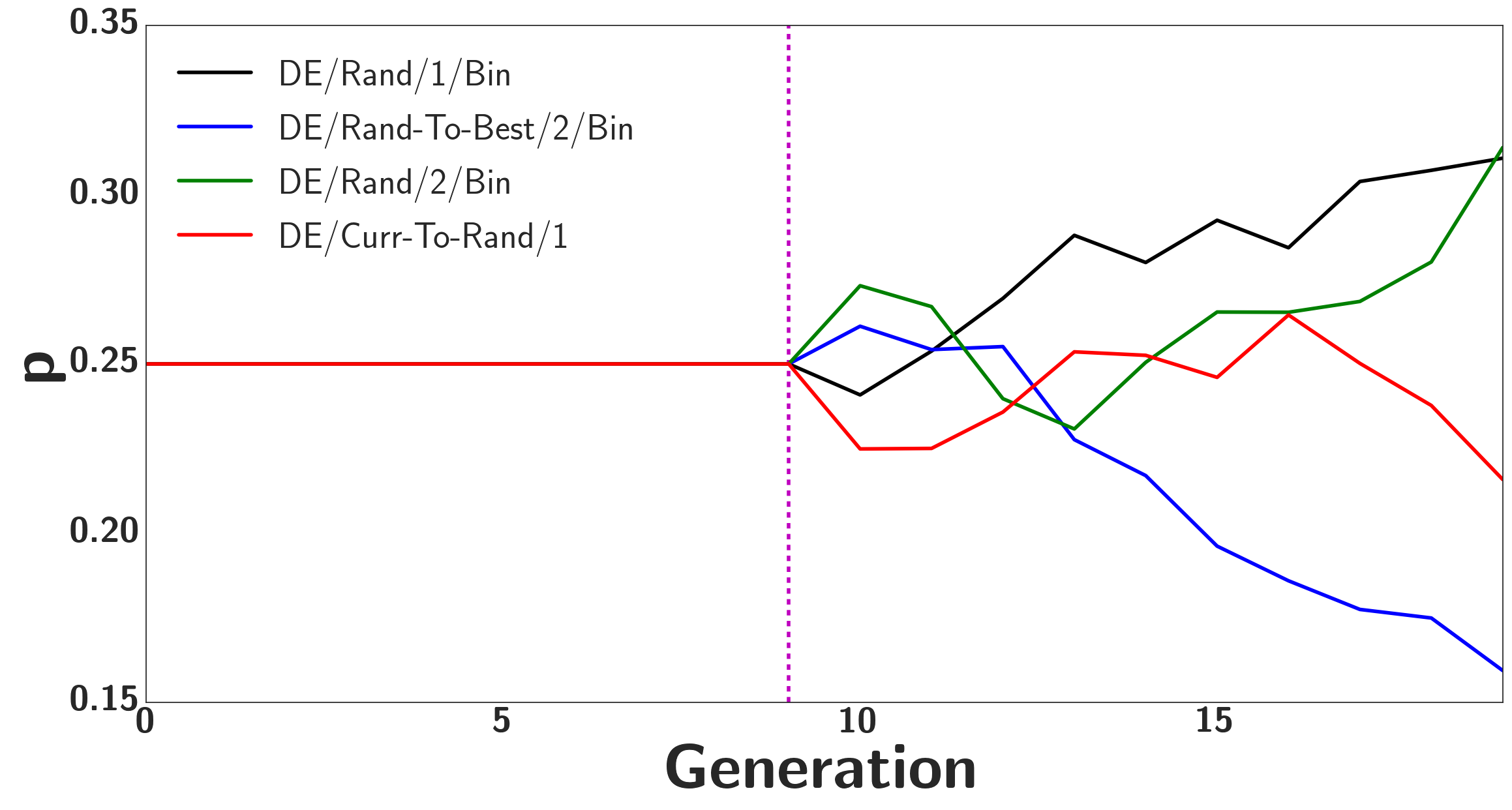}
    \caption{}\label{fig:pAVG}
  \end{subfigure}
  \caption{Averages $F_{acc}(\lambda)$, $F$, $CR$, and $p$ for the SADE population over 19 generations. The dotted vertical line indicates that the learning period has ended. Note description of the SADE algorithm in Section~\ref{sec:SADE}.}
  \label{fig:SADEALL}
\end{figure}

The SADE algorithm performed the best, in terms of fitness, out of all of the algorithms. \fig{fig:sAVG} shows the average $F_{acc}$ of the population over 19 generations. The average $F_{acc}$ converged by five iterations. The maximum $F_{acc}$ starts off at 0. This indicates that a 100\% accuracy candidate was found in the initialisation period. The maximum $F_{acc}$ then rises to 400 which is not visible as the range of the average score is -50000 to -1500.\par

The $F$ average results in \fig{fig:fAVG} are quite interesting. They start off at 1 as they are selected from $U([0,2])$ and then drop down to 0.5 as they are selected from $U([0,1])$ after the first generation.  Once the learning period has finished all of the $F$ values have converged to less than 0.1. This indicates that the $F$ values that are having the most success are small and therefore taking advantage of exploration rather than exploitation. It was unexpected that the algorithm would find a min/max within so few generations. This could be why the authors select initial $F$ from $N(0.5,0.3)$ with range [-0.4,1.4]. 

\fig{fig:crAVG} shows how the crossover probability CR for each function changes over time.  For the first nine generations, the CR values are selected from $U([0,1])$ and so the mean stays at 0.5. However, as with the $F$ mean values once the learning period is over, all of the CR values go down to less than 0.1. This means that less than 10\% of the mutations will generally take place. From a set of 11 hyper-parameters this means that probabilistically one value will change in addition to the random index that is chosen.

\begin{table}[h!tbp]
\centering
\caption{Parameters used by each model. \label{tab:PARAMS}}
\begin{tabular}{|l||c|c|c|c||}
\hline
 \textbf{Parameter} & \textbf{LGMD} &\textbf{A}&\textbf{P}&\textbf{AP}\\
\hline\hline
 $\mathbf{\tau_e (ms)}$  & 5.87&5.87&5.87&5.87 \\\hline
 $\mathbf{\tau_{iA} (ms)}$  & 3.57&3.57&3.57&3.57 \\\hline
 $\mathbf{\tau_{iB} (ms)}$ & 4.20&4.20&4.20&4.20 \\\hline
 $\mathbf{q_{eP} (pA)}$ & 1014.00&1014.00&1014.00&1014.00 \\\hline
 $\mathbf{q_{eS} (pA)}$ &4635.30&4635.30&4635.30&4635.30  \\\hline
 $\mathbf{q_{eIP} (pA)}$&84.26&84.26 & 84.26& 84.26 \\\hline
 $\mathbf{q_{eIS} (pA)}$ & 168.11&168.11&168.11&168.11 \\\hline
 $\mathbf{q_{eL} (pA)}$ & 80.00& 100.00 &80.00 &100.00  \\\hline
 $\mathbf{inhA_S (1)}$ &1.19&1.19&1.19&1.19\\\hline
 $\mathbf{inhB_S (1)}$ &1.50&1.50&1.50&1.50\\\hline
 $\mathbf{inhA_L (1)}$ &0.14&0.14&0.14&0.14 \\\hline
 $\mathbf{a (1)}$ &-&0.79&-&0.79\\\hline
 $\mathbf{b (1)}$ &-&14.51&-&14.51\\\hline
 $\mathbf{\tau_{w_{adapt}} (ms)}$ &-&30.00&-&30.00 \\\hline
 $\mathbf{\tau_{pre} (ms)}$ &-&-&1.56&1.56 \\\hline
 $\mathbf{\tau_{post (ms)}}$ &-&-&10.03&10.03\\\hline
 $\mathbf{\Delta_{pre} (1)}$ &-&-&0.031&0.031\\\hline
 $\mathbf{\Delta_{post} (1)}$ &-&-&0.027&0.027\\\hline
 $\mathbf{c (1)}$ &-&-&0.05&0.05\\\hline
\hline
\end{tabular}
\end{table} 

The probability of each function being chosen is shown in \fig{fig:pAVG}. The probabilities are fixed at 0.25 for the first 9 generations and then they vary based on their success. It is interesting to see that in spite of the $F$ and $CR$ values suggesting that the algorithm is converging on a solution, the DE/Rand-to-Best/2/Bin algorithm is the least successful. The DE/Curr-to-Rand/1 algorithm performs relatively well until about 12 generations where it tapers off. The DE/Rand/2/bin algorithm dips initially but then increases as DE/Curr-to-Rand/1 starts to drop off. The DE/Rand/1/bin remains relatively high during the entire algorithm only to be overtaken by The DE/Rand/2/bin in the last generation.

\subsection{Comparison of models}
\label{ssec:modcomp}
\tab{tab:PARAMS} shows the selected final parameters of each model. These values were all found by the SADE algorithm, due to the superior quality of its results. The (1) tag in the parameter column indicates that the variable is unit-less.

In both models with plasticity, the clamping value $c$ was set to 0.05, or 5\%.

\begin{table}[h!tbp] 
\centering
\caption{Quality metrics of the performance of different LGMD models for different simulated looming stimuli. }\label{tab:LGMDComp}
\begin{tabular}{|l|c||c|c|c|c||}\hline
\textbf{Stimulus}&\textbf{Model}&\textbf{Acc}&\textbf{Sen}&\textbf{Pre}&\textbf{Spe}\\\hline\hline
\multirow{5}{*}{\textbf{comp}}
&\textbf{LGMD}&0.90&1.00&0.83&0.80\\\cline{2-6}
&\textbf{A}&0.90&1.00&0.83&0.80\\\cline{2-6}
&\textbf{P}&0.90&1.00&0.83&0.80\\\cline{2-6}
&\textbf{AP}&0.90&1.00&0.83&0.80\\\hline
\multirow{5}{*}{\textbf{circleSlow}}
&\textbf{LGMD}&0.80&0.60&1.00&1.00\\\cline{2-6}
&\textbf{A}&0.80&0.60&1.00&1.00\\\cline{2-6}
&\textbf{P}&0.90&0.80&1.00&1.00\\\cline{2-6}
&\textbf{AP}&1.00&1.00&1.00&1.00\\\hline
\multirow{5}{*}{\textbf{circleFast}}
&\textbf{LGMD}&1.00&1.00&1.00&1.00\\\cline{2-6}
&\textbf{A}&1.00&1.00&1.00&1.00\\\cline{2-6}
&\textbf{P}&1.00&1.00&1.00&1.00\\\cline{2-6}
&\textbf{AP}&1.00&1.00&1.00&1.00\\\hline
\multirow{5}{*}{\textbf{squareSlow}}
&\textbf{LGMD}&1.00&1.00&1.00&1.00\\\cline{2-6}
&\textbf{A}&1.00&1.00&1.00&1.00\\\cline{2-6}
&\textbf{P}&1.00&1.00&1.00&1.00\\\cline{2-6}
&\textbf{AP}&1.00&1.00&1.00&1.00\\\hline
\multirow{5}{*}{\textbf{squareFast}}
&\textbf{LGMD}&1.00&1.00&1.00&1.00\\\cline{2-6}
&\textbf{A}&1.00&1.00&1.00&1.00\\\cline{2-6}
&\textbf{P}&1.00&1.00&1.00&1.00\\\cline{2-6}
&\textbf{AP}&1.00&1.00&1.00&1.00\\\hline
\end{tabular}
\end{table}

\begin{figure}[t!p]
  \centering
  \begin{subfigure}{1\columnwidth}
    \includegraphics[width=0.9\columnwidth]{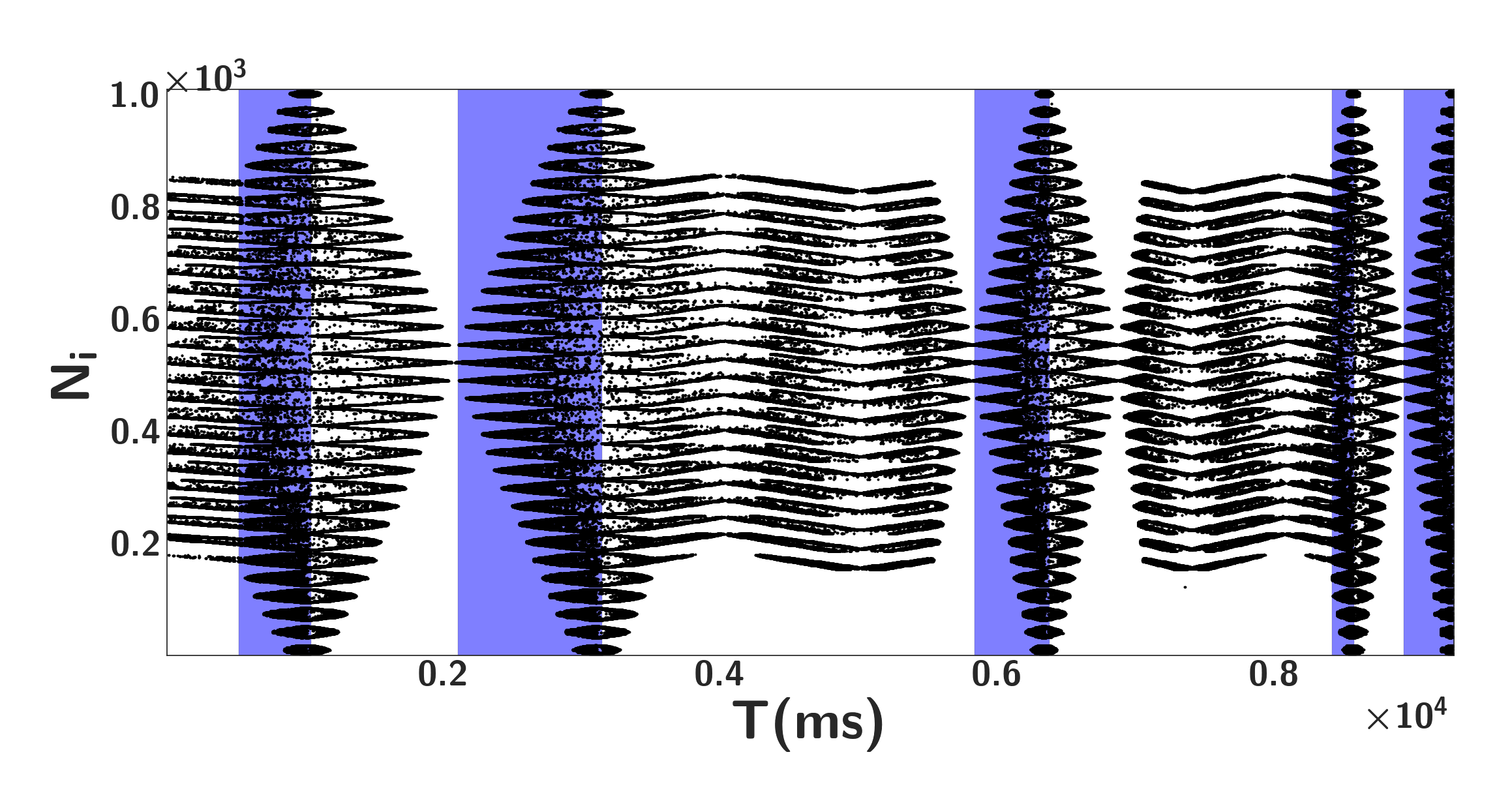}
    \caption{Filtered Composite Input (P Layer Raster Plot).}\label{fig:comP}
  \end{subfigure}
  \begin{subfigure}{1\columnwidth}
    \includegraphics[width=0.9\columnwidth]{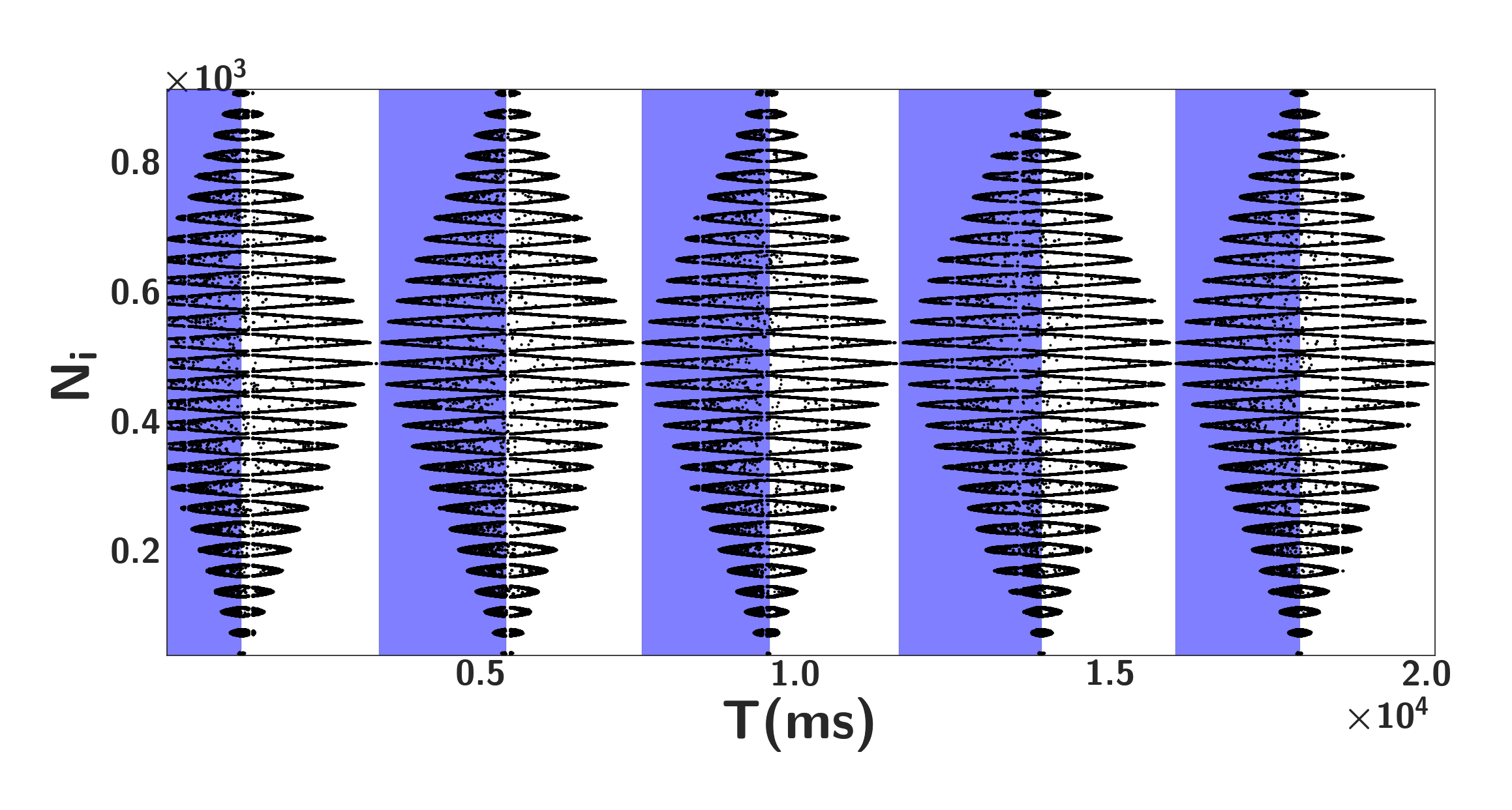}
    \caption{Filtered circleSlow Input (P Layer Raster Plot).}\label{fig:cSP}
  \end{subfigure}
  \begin{subfigure}{1\columnwidth}
    \includegraphics[width=0.9\columnwidth]{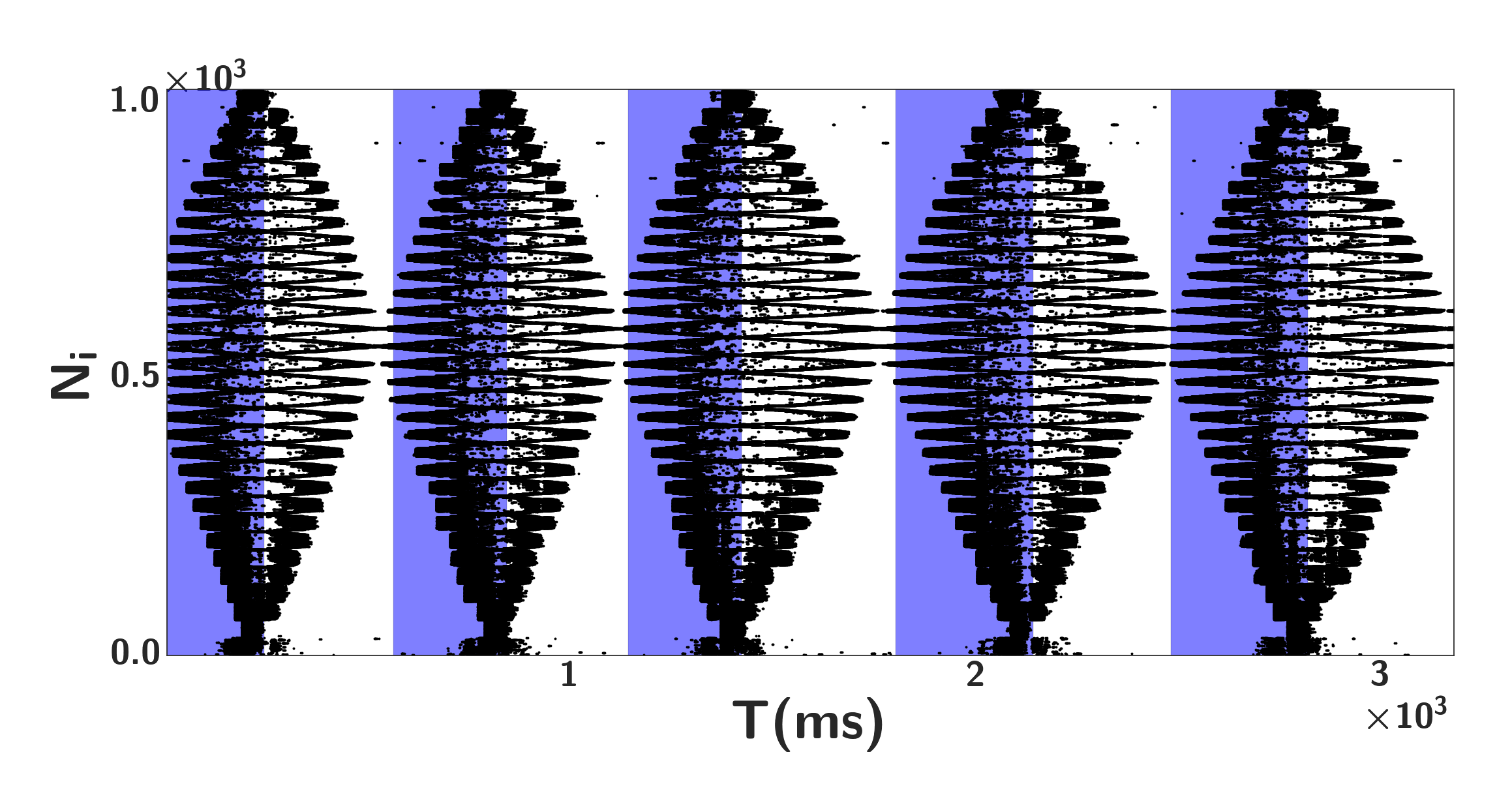}
    \caption{Filtered squareFast Input (P Layer Raster Plot).}\label{fig:sFP}
  \end{subfigure}
  \caption{The input layer for the simple stimuli.The white and coloured backgrounds indicate non-looming and looming respectively.}\label{simpleStim}
\end{figure}

As expected, all of the models have a $\tau_{iA} < \tau_{iB}$ which means that the $B$ inhibitions will persist for longer and have slower dynamics relative to the $A$ inhibitions. What is unexpected is that the $B$ inhibitions also have stronger current injection than the $A$ inhibitions. On top of this, both of the inhibitory current injections are actually stronger than the excitatory connections. Whereas the model in~\cite{yue2010reactive} with discrete dynamics had relatively low inhibitory current injections, with $inhA_S = 0.25$ and $inhB_S = 0.125$ of the excitation strength. Clearly, there is a difference between the neuron models that are used, but this is an interesting outcome nonetheless.

\begin{figure}[h!tpb]
  \centering
  \begin{subfigure}{1\columnwidth}
    \includegraphics[width=0.9\columnwidth]{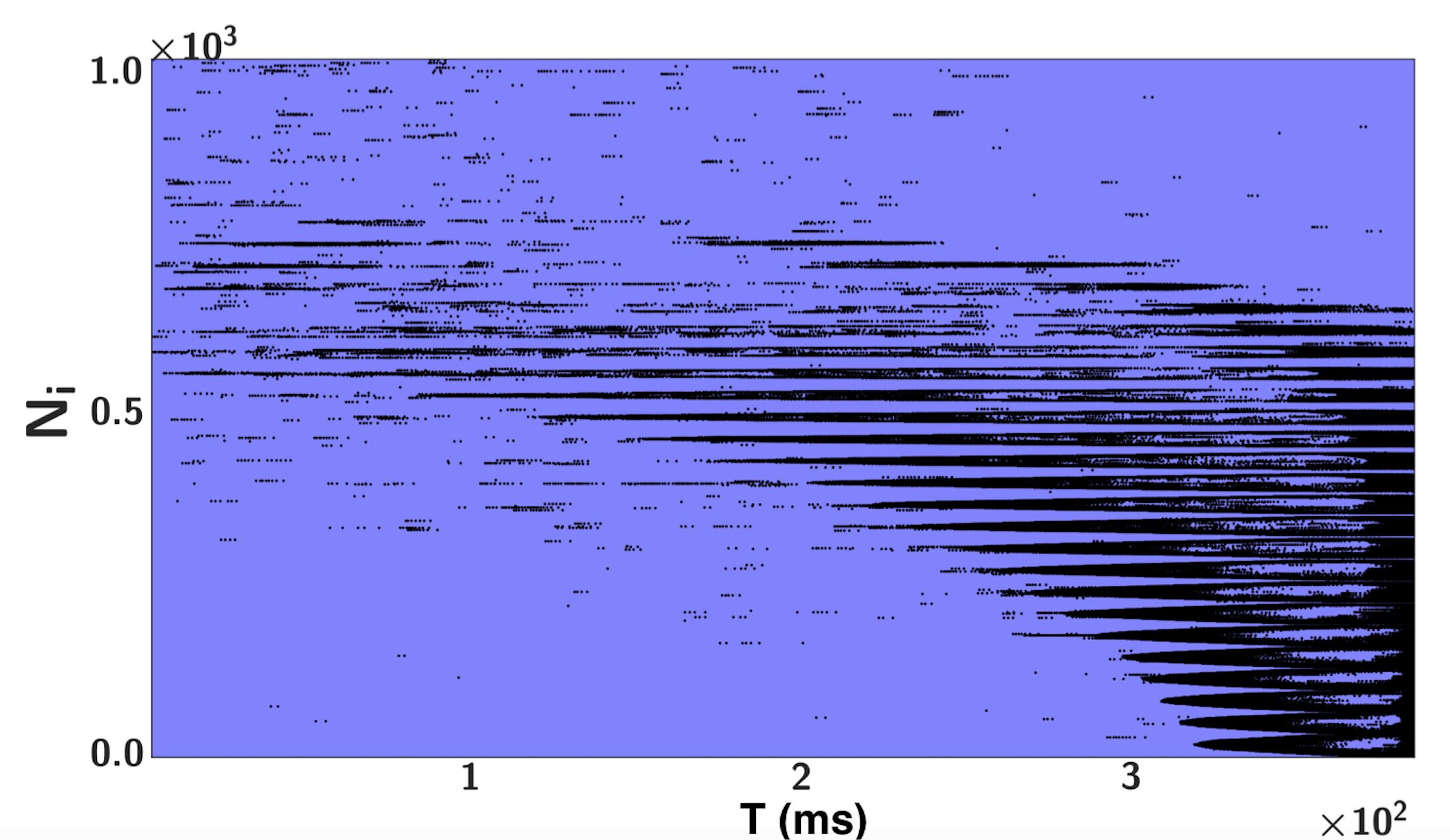}
    \caption{Filtered ballRoll2 Input (P Layer Raster Plot).}\label{fig:BRP}
  \end{subfigure}
  \begin{subfigure}{1\columnwidth}
    \includegraphics[width=0.9\columnwidth]{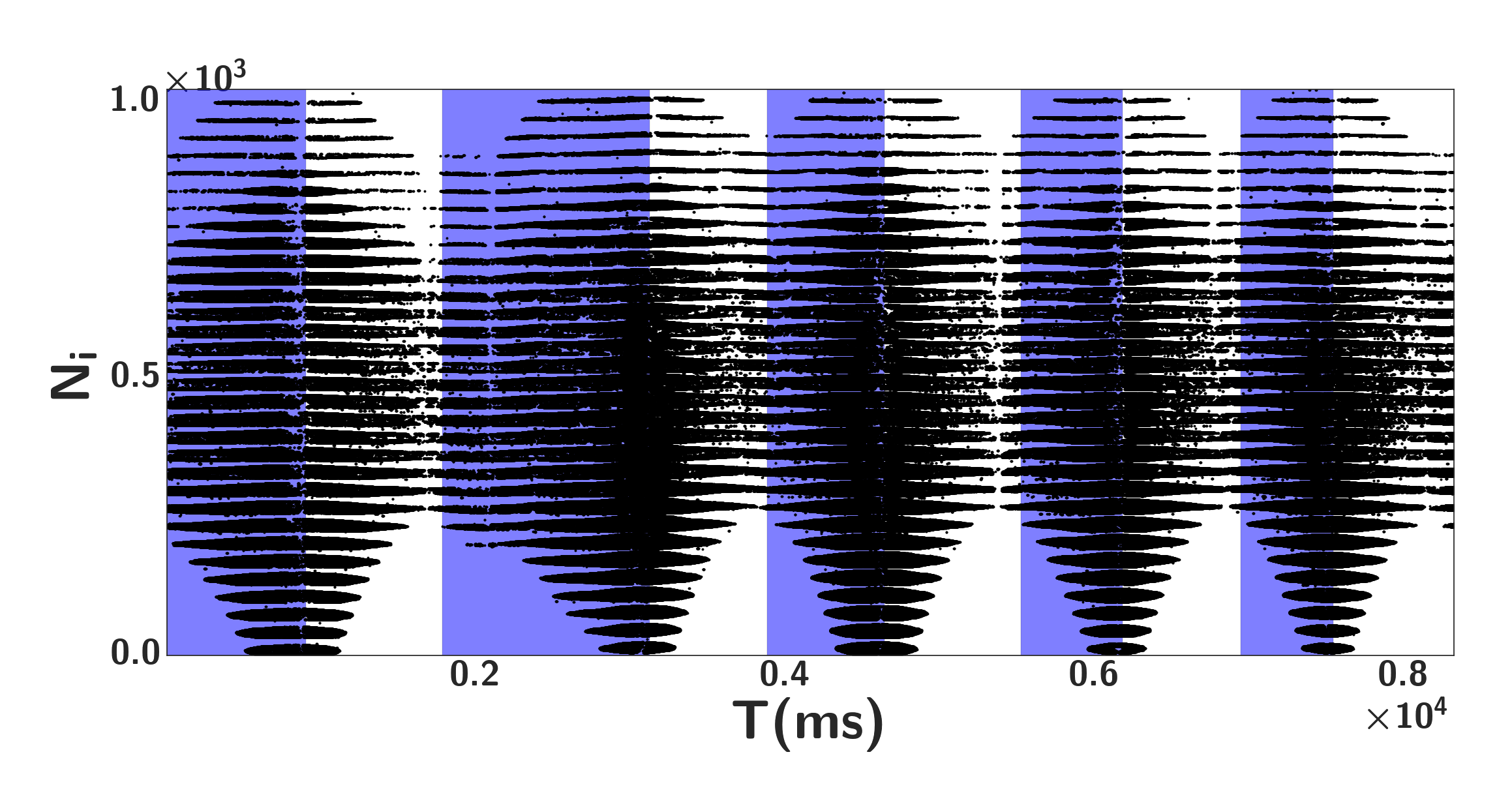}
    \caption{Filtered cupQUAV Input (P Layer Raster Plot).}\label{fig:lCP}
  \end{subfigure}
  \begin{subfigure}{1\columnwidth}
    \includegraphics[width=0.9\columnwidth]{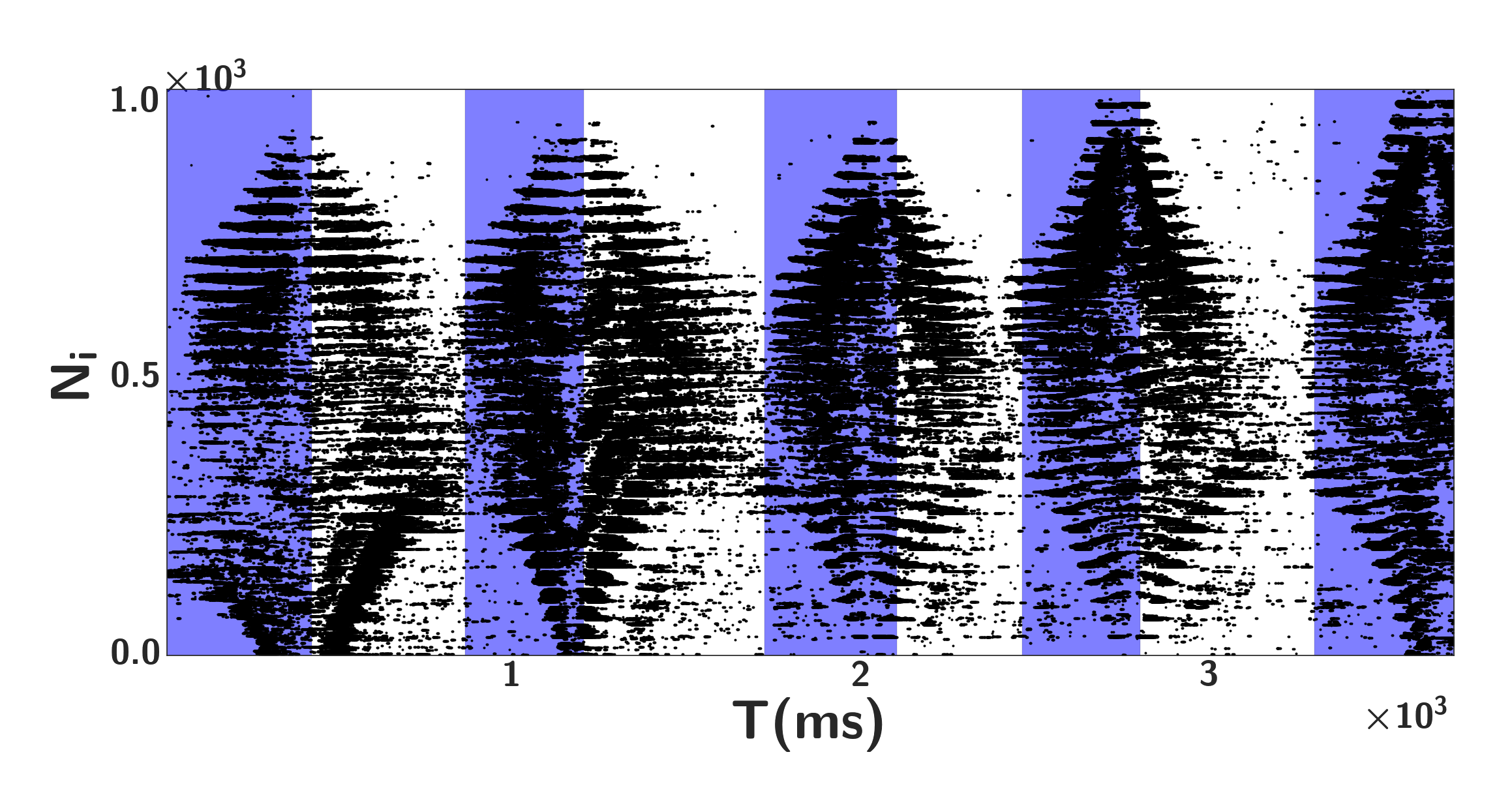}
    \caption{Filtered Hand Input (P Layer Raster Plot).}\label{fig:hP}
  \end{subfigure}
  \caption{Complex real stimuli. The white and coloured backgrounds indicate non-looming and looming respectively.}
\end{figure}

\tab{tab:LGMDComp} shows the accuracy, sensitivity, precision, and specificity for each LGMD model for a given simple stimulus. The stimuli can be described as follows: 

\begin{description}[topsep=0.2ex]
\item[\textbf{composite:}] A standard test bench stimulus that consists of a black circle on a white background that translates and looms at increasing speeds. \fig{fig:comP} shows the composite input. 
\item[\textbf{circleFast/Slow:}] A purely looming black circle on white background at high or low speeds. Collected on hovering QUAV. \fig{fig:cSP} shows the circleFast/Slow stimulus. 
\par
\item[\textbf{squareFast/Slow:}] A purely looming black square on a white background at high/low speeds. \fig{fig:sFP} shows the squareFast/Slow stimulus.
\end{description}

The results in \tab{tab:LGMDComp} show that the models performed well ($Accuracy\geq 0.8$) on most of the stimuli. LGMD and \textbf{A} perform poorly on the circleSlow test, missing two out of five of the looming stimuli. \textbf{P} misses one looming stimulus, and \textbf{AP} detects all stimuli accurately. The plasticity increases the weights of important connections and the adaptation filters out over excited neurons.

These results show that the models are capable of detecting looming stimuli of varying speeds and of differentiating between translation and looming stimuli for the most part. \textbf{AP} scored 100\% in every test besides the composite stimulus where it misclassified the first short translation as a loom. It is likely that this is due to the network not starting in its resting/equilibrium state. Inspecting the output trace of \textbf{LGMD} on the \textbf{composite} stimulus it takes the model \~150$ms$ to stop spiking after the looming phase has finished. 

After performing the simulated experiments of computer generated shapes, real objects moving towards and away from the camera were recorded. These stimuli can be described as:
\begin{description}
\item[\textbf{ballRoll[1-3]:}] Three different runs of a white ball rolling towards the camera on a black platform at different angles and speeds. This is a purely looming stimulus. \fig{fig:BRP} shows one of the three ball rolls.
\item[\textbf{cupQUAV:} ] A QUAV flying towards a cup suspended in front of it with a white wall behind it. This is a self stimulus.\fig{fig:lCP} shows the QUAV cup stimulus.
\item[\textbf{Hand:} ] A Hand moving towards and away from the hovering QUAV.  \fig{fig:hP} shows the looming hand stimulus. 
\end{description}\par

\fig{fig:BRP}, \fig{fig:lCP}, and \fig{fig:hP} show that the real stimuli tend to have more noise and do not adhere to a strong pattern when compared to \fig{fig:comP}, \fig{fig:cSP}, and \fig{fig:sFP}. \tab{tab:LGMDComp2} shows that the models do not perform as well on real world stimuli. ballRoll[1-3] is the simplest real stimulus, and as such \textbf{P} and \textbf{AP} achieved full accuracy. LGMD and \textbf{A} missed one roll.

\begin{table}[h!tbp]
\centering
\caption{Quality metrics of the performance of different LGMD models for different real looming stimuli}\label{tab:LGMDComp2}
\begin{tabular}{|l|c||c|c|c|c||}\hline
\textbf{Stim}&\textbf{Model}&\textbf{Acc}&\textbf{Sen}&\textbf{Pre}&\textbf{Spe}\\\hline\hline
\multirow{5}{*}{\textbf{ballRoll[1-3]}}
&\textbf{LGMD}&0.66&0.66&1.00&0.00\\\cline{2-6}
&\textbf{A}&0.66&0.66&1.00&0.00\\\cline{2-6}
&\textbf{P}&1.00&1.00&1.00&0.00\\\cline{2-6}
&\textbf{AP}&1.00&1.00&1.00&0.00\\\hline
\multirow{5}{*}{\textbf{cupQUAV}}
&\textbf{LGMD}&0.70&1.00&0.62&0.40\\\cline{2-6}
&\textbf{A}&0.70&1.00&0.62&0.40\\\cline{2-6}
&\textbf{P}&0.70&1.00&0.62&0.40\\\cline{2-6}
&\textbf{AP}&0.80&1.00&0.71&0.60\\\hline
\multirow{5}{*}{\textbf{hand}}
&\textbf{LGMD}&0.50&1.00&0.50&0.00\\\cline{2-6}
&\textbf{A}&0.50&1.00&0.50&0.00\\\cline{2-6}
&\textbf{P}&0.50&1.00&0.50&0.00\\\cline{2-6}
&\textbf{AP}&0.50&1.00&0.50&0.00\\\hline
\end{tabular}
\end{table}

Surprisingly good results come from the cupQUAV stimulus: 70\% accuracy for all models except for \textbf{AP}, which had 80\%. It is worth noting that \textbf{AP} performed consistently well when compared with the other models.

The real world stimuli tended to contain more activations due to the irregularity of the shapes and increased noise. The possibility of detecting the hand by stochastically dropping pixel-events, was investigated. Dropping 50\% of the DVS events and re-optimising the network gave 100\% accuracy for the hand and cupQuad stimulus. However, in doing this, the network was no longer robust to the speed changes in the composite benchmark test. Indeed, even using all of the pixels, the network could be optimised to work on the real world stimuli, but would no longer be as accurate for inputs that contained less activations. The inhibition values went up and the gain values went down, meaning the network struggled to spike on stimuli that weren't noisy or event heavy. Some sort of additional pre-filtering could be useful in getting the looming network to be fully robust in all situations, such as a sub-sampling filter that allows less than some maximum number of input pixels to be active at a given time stepls to be active at a given time step. 

\subsubsection{The effect of changing $c$ on plasticity}
\label{ssec:plasC}

\begin{figure}[h!tbp]
\centering
  \begin{subfigure}{0.9\columnwidth}
    \includegraphics[width=1\columnwidth]{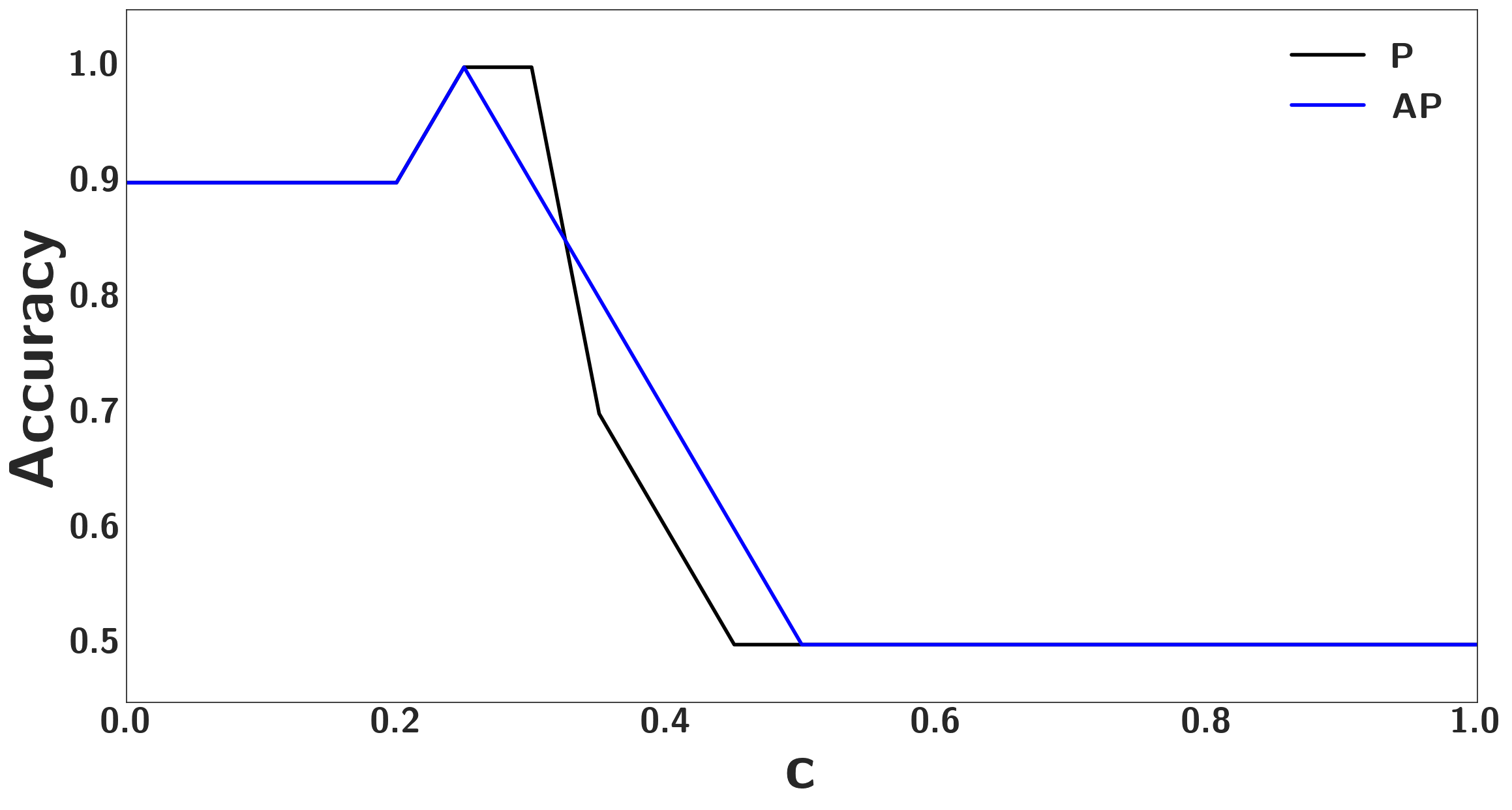}
    \caption{Effect of changing the $c$ clamping value on the learning weight $w$ for the composite stimulus}\label{fig:pCom}
  \end{subfigure} 
  \begin{subfigure}{0.9\columnwidth}
    \includegraphics[width=1\columnwidth]{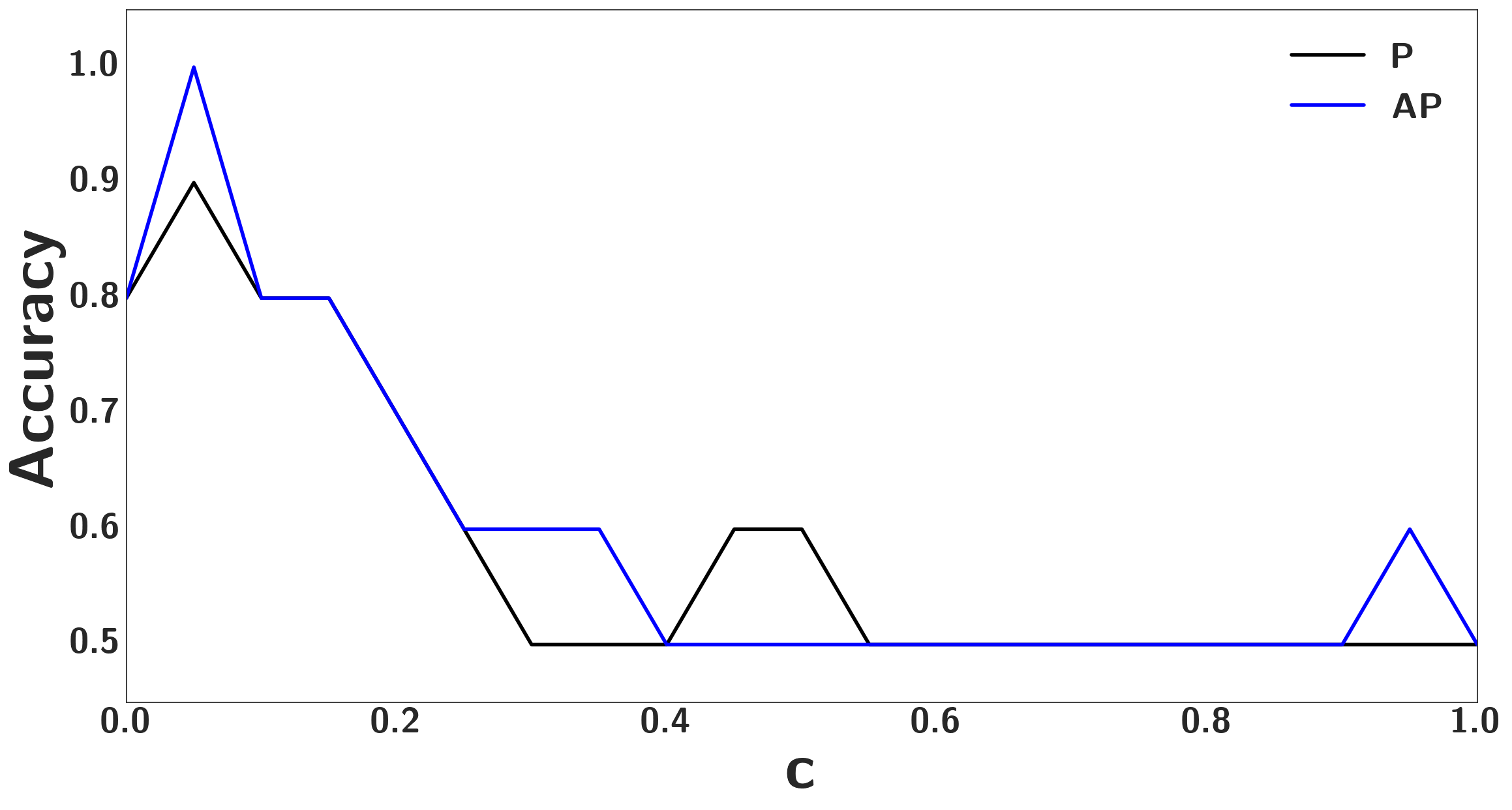}
    \caption{Effect of changing the $c$ clamping value on the learning weight $w$ for the circleSlow stimulus}\label{fig:psloC}
  \end{subfigure} 
  \begin{subfigure}{0.9\columnwidth}
    \centering
    \includegraphics[width=1\columnwidth]{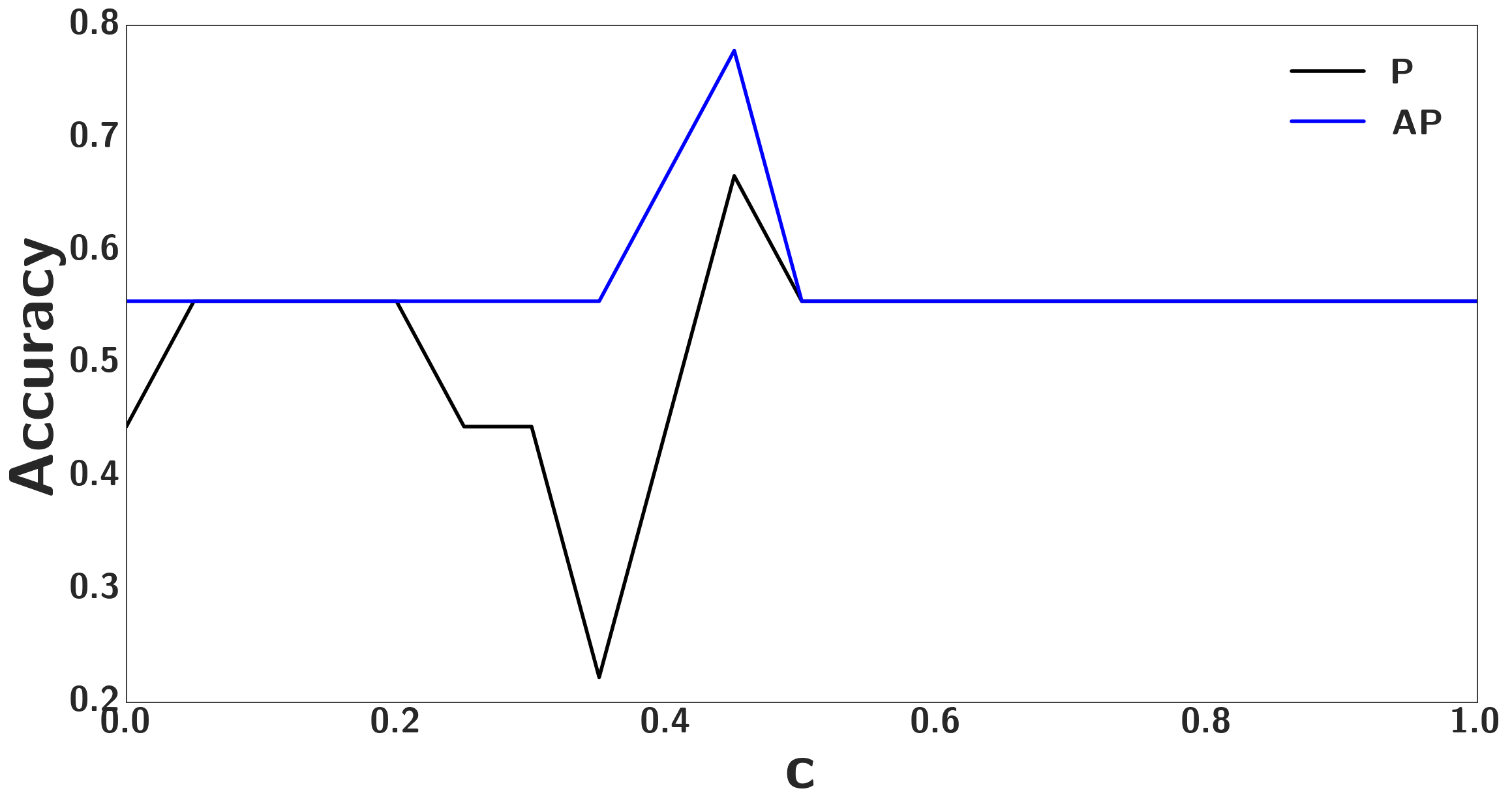}
    \caption{Effect of changing the $c$ clamping value on the learning weight $w$ for the hand stimulus}\label{fig:pH}
  \end{subfigure}
  \caption{The effect of changing the clamping value on various stimuli}
\end{figure}

\fig{fig:pCom}, \fig{fig:psloC}, and \fig{fig:pH} show how changing the bounds of the plasticity clamping changes the LGMD (\textbf{P} model) accuracy for the composite, cicleSlow, and hand stimuli respectively.

Interestingly, for the two simulated stimuli increasing the clamping to beyond 25\% caused the accuracy to drop to 50\%. The sensitivity dropped to 0\% indicating that it was no longer detecting looms and that the synaptic weights were no longer causing the LGMD neuron to fire.

Increasing the clamping to 45\% increases the accuracy for both the \textbf{P} and \textbf{AP} models on the hand stimulus. This shows that plasticity is a double edged sword that can both improve and degrade the performance of the model. In the simulated stimuli lower clamping, $c=0.1-0.2$, tended to perform better but in the hand stimulus larger, $c=0.5$, gave better results. This could suggest that too much plasticity can cause the weights to deviate too far from their good values on well formed stimuli but help to reduce noise in real world stimuli. Knowledge about the nature of your input can help to determine what level of plasticity you require. In all simulated and real cases except for the hand stimulus, a small contribution of plasticity improved the performance.

\section{Conclusions}
We implemented a neuromorphic model of the locust LGMD network using recordings from a UAV equipped with a DVS sensor as inputs. The neuromorphic LGMDNN was capable of differentiating between looming and non-looming stimuli. It was capable of detecting the black and white simple stimuli correctly regardless of speed and shape. Real-world stimuli performed relatively well using the parameters found by the optimiser for synthesised stimuli. However, when re-optimised, the real-world stimuli performed comparably to the synthesised stimuli. This was mainly because real-world stimuli tend to contain a higher number of luminance changes and therefore the magnitude parameters needed to be reduced. 

We showed that BO, DE, and SADE are capable of finding parameter values that give the desired performance in the LGMDNN model. It can be seen that SADE statistically significantly outperformed DE on all metrics besides specificity and the number of evaluations,  although the only metrics that formed part of the objective function were fitness and accuracy. Once a suitable objective function was found that accurately described the desired output of the LGMDNN, BO, DE and SADE outperformed hand-crafted attempts, but a uniform random search also performed well. The algorithms were able to achieve 100\% accuracy on  black and white simple stimuli of varying shapes and speeds. SADE performed well in this task and we have shown that it is suitable for the optimisation of a multi-layered LGMD spiking neural network. This could save time when developing biologically plausible SNNs in related applications.

We have also studied effect of synaptic plasticity and neuronal spike-frequency adaptation on the performance of the LGMDNN, using the most successful parameter optimisation method. Our conclusion is that plasticity plays an important role in increasing (and decreasing) performance, depending on how its parameters are selected.

In the future, we plan to apply the optimisation algorithms directly to tuning the neuromorphic processors implementation of the model, with the end goal being a closed loop control system on a UAV\@. Showing that the optimisation approach proposed is effective for selecting parameters directly on closed-loop neuromorphic hardware set-ups will greatly increase their usability.

%

\ifCLASSOPTIONcompsoc
  \section*{Acknowledgments}
\else
  \section*{Acknowledgment}
\fi

We are grateful to Prof. Claire Rind, who provided valuable comments and feedback on the definition of the neuromorphic model, and acknowledge the CapoCaccia Cognitive Neuromorphic Engineering workshop, where these discussions and model developments took place.

We would also like to thank iniLabs for use of the DVS sensor and the Institute of Neuroinformatics (INI), University of Zurich and ETH Zurich for its neuromorphic processor developments. Part of this work was funded by the EU ERC Grant ``NeuroAgents'' (724295) and SNSF Ambizione grant (PZOOP2\_168183).

We would also like to thank Associate Professor Marcus Gallagher who read the paper and provided some insight as to why the uniform random walk may outperform other optimisers.
\ifCLASSOPTIONcaptionsoff
  \newpage
\fi



%
\bibliographystyle{IEEEtran}
\bibliography{TNNLS,biblio}

%



\begin{IEEEbiography}[{\includegraphics[width=1in,height=1.25in,clip,keepaspectratio]{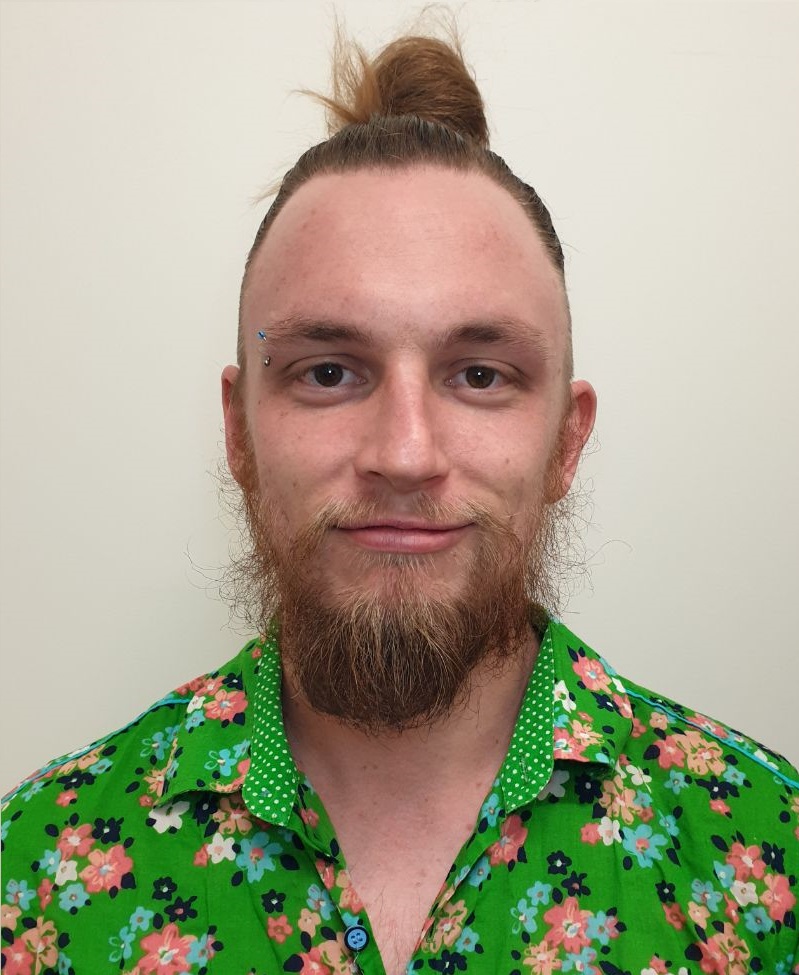}}]{Llewyn Salt}
is a PhD candidate in Reinforcement Learning at the University of Queensland, Australia. His research is focused on utilising goals in continuous control problems in reinforcement learning. 

He received a bachelor's (first class honours) and master's degree in mechatronic engineering from the University of Queensland. The work in this paper was undertaken at the Institute of Neuroinformatics at the University of Zurich and ETH Zurich as part of his master's degree program.  
\end{IEEEbiography}
\begin{IEEEbiography}[{\includegraphics[width=1in,height=1.25in,clip,keepaspectratio]{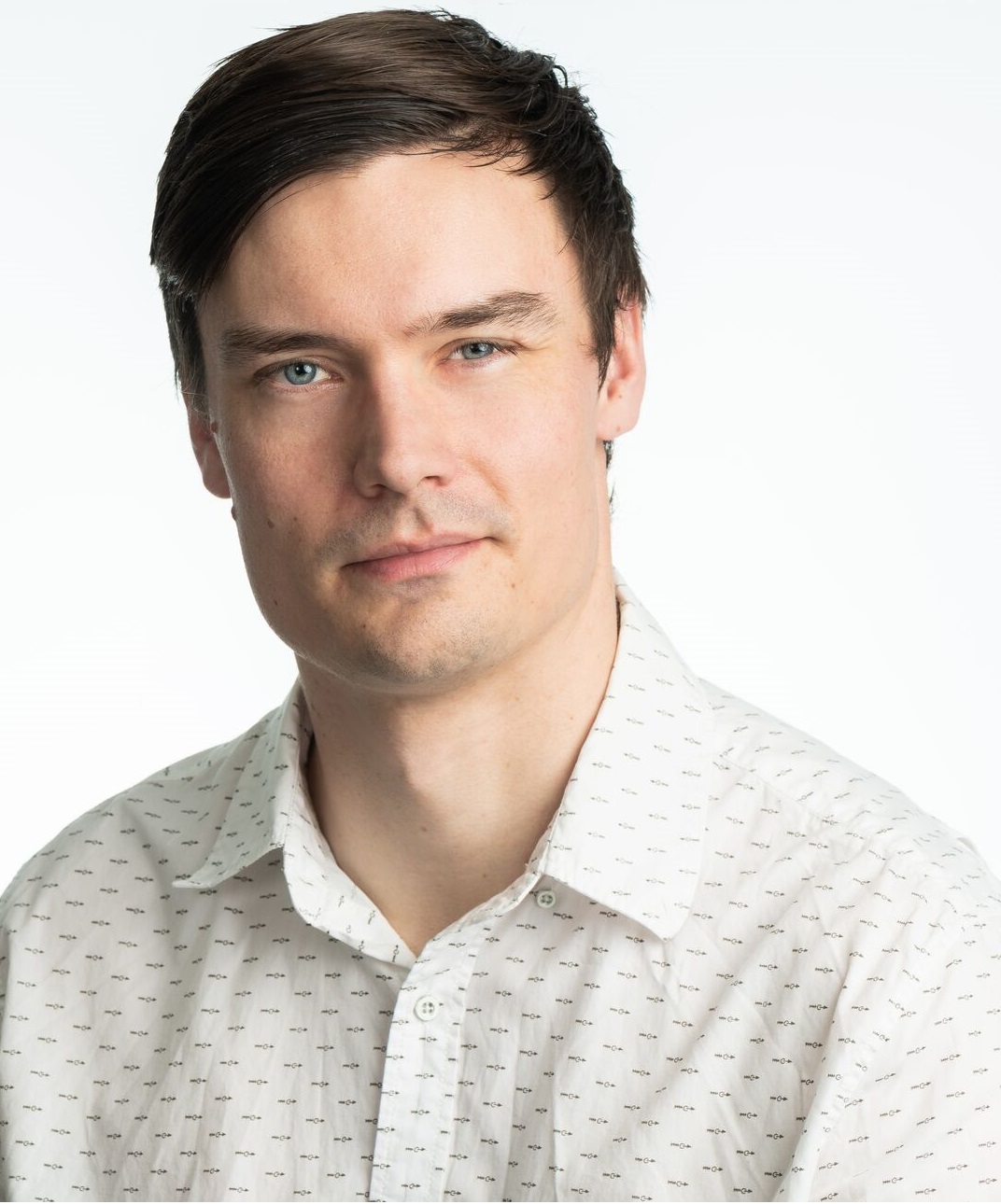}}]{David Howard}
David is a Senior Research Scientist at CSIRO, Australia's national science body.  He works and leads multiple projects at the intersection of robotics, evolutionary machine learning, and robotic materials.  His interests include nature-inspired algorithms, learned autonomy, soft robotics, the reality gap, and evolution of form.  His work has been featured in local and national media, TechXplore, and Wired.

He received his BSc in Computing from the University of Leeds in 2005, and completed a MSc in Cognitive Systems at the same institution in 2006.  In 2011 he received his PhD from the University of the West of England.  He is a member of the IEEE and ACM, and an avid proponent of education, STEM, and outreach activities.
\end{IEEEbiography}
\begin{IEEEbiography}[{\includegraphics[width=1in,height=1.25in,clip,keepaspectratio]{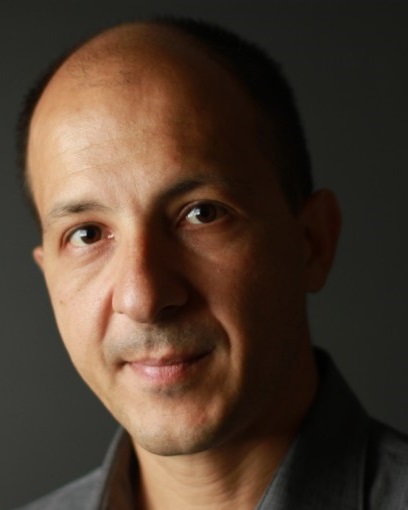}}]{Giacomo Indiveri}
Giacomo Indiveri is a Professor at the University of Zurich and ETH Zurich, Switzerland. He obtained an M.Sc. degree in Electrical Engineering and a Ph.D. degree in Computer Science from the University of Genoa, Italy. Indiveri was a post-doctoral research fellow in the Division of Biology at the California Institute of Technology (Caltech) and at the Institute of Neuroinformatics of the University of Zurich and ETH Zurich.  He was awarded three ERC grants and is an IEEE Senior member. His research interests lie in the study of real and artificial neural processing systems, and in the hardware implementation of neuromorphic cognitive systems, using full custom analog and digital VLSI technology. 
\end{IEEEbiography}
\begin{IEEEbiography}[{\includegraphics[width=1in,height=1.25in,clip,keepaspectratio]{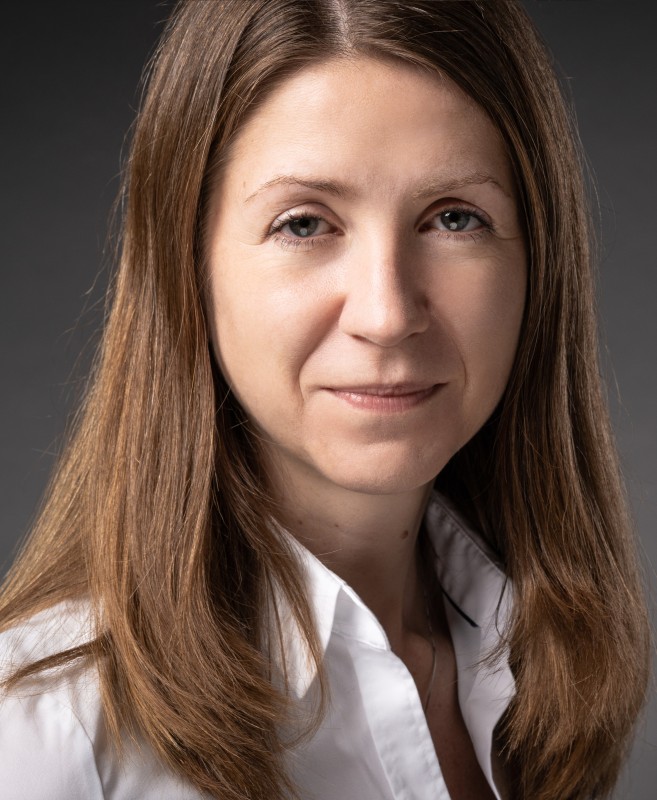}}]{Yulia Sandamirskaya}
is a Group Leader in the Institute of Neuroinformatics (INI) at the University of Zurich and ETH Zurich. Her group “Neuromorphic Cognitive Robots” develops neuro-dynamic architectures for embodied cognitive agents. In particular, she studies memory formation, motor control, and autonomous learning in spiking and continuous neural networks, realised in neuromorphic hardware interfaced to robotic sensors and motors. She has a degree in Physics from the Belarussian State University in Minsk, Belarus and Dr. rer. nat. from the Institute for Neural Computation in Bochum, Germany. She is the chair of EUCOG — the European Society for Cognitive Systems and the coordinator of the NEUROTECH project that supports and develops the neuromorphic computing community in Europe.
\end{IEEEbiography}
\end{document}